\documentclass[preprint,12pt,3p]{elsarticle}
\usepackage[T1]{fontenc}
\usepackage{times}
\usepackage{graphics,graphicx,subcaption}
\usepackage{enumerate,epsfig}
\usepackage{multirow,afterpage,rotating}
\usepackage{amsfonts}
\usepackage{amsmath,amssymb}
\usepackage{caption}
\usepackage{soul}
\usepackage{url}
\usepackage{inputenc}
\usepackage[misc,geometry]{ifsym}
\usepackage[colorlinks=true]{hyperref}
\usepackage{dcolumn}
\usepackage{wrapfig,dsfont}
\usepackage{mathtools}
\usepackage{color}
\usepackage{tikz}
\usepackage{float}
\usepackage[linesnumbered,ruled,vlined]{algorithm2e}
\usepackage{array}
\usepackage{longtable}

\newcolumntype{C}[1]{>{\centering\arraybackslash}p{#1}}

\DeclareCaptionFont{red}{\color{red}}

\newcommand{\dashedline}[1][magenta]{\raisebox{2pt}{\tikz{\draw[-,#1,dashed,line width = 1.5pt](0,0) -- (5mm,0);}}}
\newcommand{\solidline}[1][magenta]{\raisebox{2pt}{\tikz{\draw[-,#1,solid,line width = 1.5pt](0,0) -- (5mm,0);}}}

\definecolor{henna}{rgb}{0.8,0.8,0.1}
\definecolor{skyblue}{rgb}{0.3010,0.7450,0.9330}
\definecolor{purple}{rgb}{0.4940,0.1840,0.5560}
\definecolor{darkblue}{rgb}{0,0.4470,0.7410}
\definecolor{darkgreen}{rgb}{0.4660,0.6740,0.1880}
\definecolor{orange}{rgb}{0.8500,0.3250,0.0980}
\definecolor{lightorange}{rgb}{0.9290,0.6940,0.1250}
\definecolor{maroon}{rgb}{0.6350,0.0780,0.1840}

\biboptions{authoryear}

\hypersetup{urlcolor=blue, citecolor=red}

\newlength{\defbaselineskip}
\newcommand{\setlinespacing}[1]%
           {\setlength{\baselineskip}{#1 \defbaselineskip}}

\journal{Expert Systems with Applications}

\begin{document}

\begin{frontmatter}

\title{Multistep traffic speed prediction: A deep learning based approach using latent space mapping considering spatio-temporal dependencies}

\author[label1]{Shatrughan Modi\corref{cor1}}
\address[label1]{Computer Science and Engineering Department, Thapar Institute of Engineering and Technology, Patiala - 147004, Punjab, India}

\cortext[cor1]{corresponding author}
\ead{shatrughanmodi@gmail.com}

\author[label1]{Jhilik Bhattacharya}
\ead{jhilik@thapar.edu}

\author[label5]{Prasenjit Basak}
\address[label5]{Electrical and Instrumentation Engineering Department, Thapar Institute of Engineering and Technology, Patiala - 147004, Punjab, India}
\ead{prasenjit@thapar.edu}

\begin{abstract}
     Traffic management in a city has become a major problem due to the increasing number of vehicles on roads. Intelligent Transportation System (ITS) can help the city traffic managers to tackle the problem by providing accurate traffic forecasts. For this, ITS requires a reliable traffic prediction algorithm that can provide accurate traffic prediction at multiple time steps based on past and current traffic data. In recent years, a number of different methods for traffic prediction have been proposed which have proved their effectiveness in terms of accuracy. However, most of these methods have either considered spatial information or temporal information only and overlooked the effect of other. In this paper, to address the above problem a deep learning based approach has been developed using both the spatial and temporal dependencies. To consider spatio-temporal dependencies, nearby road sensors at a particular instant are selected based on the attributes like traffic similarity and distance. Two pre-trained deep auto-encoders were cross-connected using the concept of latent space mapping and the resultant model was trained using the traffic data from the selected nearby sensors as input. The proposed deep learning based approach was trained using the real-world traffic data collected from loop detector sensors installed on different highways of Los Angeles and Bay Area. The traffic data is freely available from the web portal of the California Department of Transportation Performance Measurement System (PeMS). The effectiveness of the proposed approach was verified by comparing it with a number of machine/deep learning approaches. It has been found that the proposed approach provides accurate traffic prediction results even for 60-min ahead prediction with least error than other techniques.
\end{abstract}

\end{frontmatter}

\section{Introduction}\label{Sec:Introduction}
The use of automobile vehicles is increasing at a very rapid rate which creates several social problems like traffic congestion, overconsumption of energy, traffic accidents, and high carbon emission (\cite{Sang2016}). To solve these problems Intelligent Transportation System (ITS) (\cite{Lin2017}) has been considered as a promising solution. For example, ITS can provide travelers with real-time information and can suggest optimal traveling routes based on future traffic predictions. Other than this, based on the current traffic speed and future traffic change trend ITS can help in optimizing the traffic signal timings which can ensure smooth traffic movement and hence can help in reducing traffic congestion. So, the successful implementation of ITS requires accurate and real-time access to traffic status information. This further makes accurate traffic prediction an important task and has attracted a large number of researchers and it has become a hot research topic over the past few decades. There are many techniques developed by different researchers to accurately predict the traffic state. These techniques can be categorized into three main categories: statistical techniques, shallow machine learning techniques, and deep learning techniques.

Statistical techniques mainly use time series analysis and based on the previously observed values predict the future values. One of the most commonly used statistical techniques used for traffic prediction is Auto-Regressive Integrated Moving Average (ARIMA) (\cite{Williams2003}). Multiple variations of ARIMA (like Seasonal ARIMA, and Vector ARIMA, etc.) has also been proposed by a number of researchers (\cite{Williams2001, Chandra2009, Chen2011, Kumar2015}) for traffic prediction. Other than this, statistical models like the Kalman filter have also been used in many studies for traffic prediction (\cite{Xie2007, Ojeda2013, GUO201450}). However, simple time series models work mainly on the assumption of stationary data i.e. the mean and variance of the data does not change, which is not consistent with the traffic data. Also, as the future predictions greatly depend on the previously observed values the error can propagate to multiple steps in multistep prediction. So, these simple time series models lack the capability to satisfy the high accuracy requirement for the successful application of ITS.

Several machine learning algorithms have also been proposed for traffic forecasting and these algorithms have shown promising improvement over statistical techniques. For instance, different researchers developed a number of approaches using the Artificial Neural Network (ANN) (\cite{Khotanzad, Csikos2015, Sharma2018}), Support Vector Machine (SVM) (\cite{Yang2010, Duan2018}), k-Nearest Neighbor (kNN) (\cite{Yu2016, Cai2016}), Bayesian network (\cite{Sun2006, Castillo2008}), XGBoost (\cite{8516114, Mei2018}) and random forest (\cite{Zarei2013, Liu2017}) for traffic prediction problem. Although the machine learning models have shown performance improvement over statistical techniques, these techniques rely highly on manually selected features and these features vary from problem to problem. Also, there are no standard guidelines available for selecting features. Other than this, the above-mentioned machine learning models have shallow architectures which limit their capability to extract and understand the highly complex and dynamic patterns from large historical traffic data. So, without proper features set and due to their shallow architectures, the machine learning techniques may not be best suitable for this complicated traffic prediction task.

With the advancement of technology, the capability of hardware to process a large amount of data has increased dramatically. Hence, deep learning techniques have gained popularity in the field of computer science and have also been successfully applied in the transportation sector (\cite{Modi2020,Modi2020b}). Table~\ref{Tab:LiteratureComparison} presents the comparison of various deep learning based approaches to predict future traffic pattern based on different parameters. \cite{s19102229} used stacked auto-encoders for predicting traffic congestion in the transport network. They have used time series of snapshots of network traffic congestion maps as input to the model. The stacked auto-encoder model predicted traffic congestion very efficiently but did not consider the effect of other parameters like traffic flow, occupancy, speed, and volume, etc. \cite{s17040818} proposed the usage of Convolutional Neural Network (CNN) for traffic speed prediction. Traffic speed data from various sensors was converted into images before giving them as input to the CNN. Here, they have considered the traffic speed data from nearby sensors located on the same road only and hence did not consider the effect of traffic from neighboring roads. Similarly, \cite{Dai2019} used CNN to forecast traffic flow by understanding the historical traffic flow pattern from traffic data of district 4 of California, USA. A Deep Belief Network (DBN) based approach has been developed in \cite{LI20191} for traffic flow prediction. The DBN was optimized using the algorithm of multi-objective particle swarm optimization which increases the time for forecasting. Also, the model was trained and tested on a dataset obtained from only nine detectors installed on a small section of a highway in Wisconsin, so the scalability of the approach has not been tested. Other than these, to consider the temporal dependency of traffic data, approaches using Long Short Term Memory (LSTM) has also been developed. For instance, \cite{Cui2018,s19183836} and \cite{Chen2021} developed LSTM based approach for traffic prediction. Some researchers have developed graph based approaches to capture spatial traffic dependency. For instance, \cite{Learning2018, Zhao2020} and \cite{Li2021a} have used Graph Convolutional Network (GCN) to predict the freeway traffic. Some researchers have developed techniques by focusing on different traffic dependencies like temporal or spatial dependencies. For example, \cite{Zheng2020,Yu2021a} and \cite{Yin2021} used graph network along with different attention mechanisms to focus on particular dependency and predict the near future traffic.

Several hybrid approaches have also been developed to predict future traffic. For instance, \cite{DeMedrano2020} combined CNN and auto-encoder to predict traffic flow based on historical weather and traffic data. Similarly, \cite{Lu2021} used ARIMA with LSTM, \cite{Du2020} used CNN with Gated Recurrent Units (GRU), \cite{lv2018lc} integrated Recurrent Neural Network (RNN) with CNN and \cite{Peng2020} combined Genetic Algorithm (GA) and back-propagation Neural Network (NN) with wavelet decomposition to learn and predict traffic pattern. It has been found that with the increase in prediction time horizon i.e. when predicting for distant time instant in future like 60-min, the prediction accuracy decreases (\cite{Li2017, Learning2018}). Also, the prediction accuracy decreases for traffic prediction in peak hours for most of the deep learning based algorithms (\cite{Kumar2015, Sharma2018}). The main reason for this is that these studies have either considered spatial dependency or temporal dependency. Also, most of the studies while considering spatial dependency, only considered the traffic information from upstream or downstream sensors and have completely neglected the effect of traffic from other neighboring roads \cite{6033614, s17040818}. Although there are some short-comings, the deep learning algorithms have shown promising results than shallow machine learning and statistical algorithms which motivated the authors for their use.

\begin{center}
\footnotesize
\begin{longtable}{|C{9em}|C{7em}|C{4.5em}|C{7em}|C{4em}|c|C{2.5em}|C{2.5em}|}

  \caption{Comparison of different deep learning based techniques for traffic prediction presented in the literature}
  \label{Tab:LiteratureComparison}\\

    \hline
    \multirow{2}[4]{*}{Ref} & \multirow{2}[4]{*}{Technique} & \multirow{2}[4]{4.5em}{Prediction Parameter} & \multirow{2}[4]{*}{Dataset} & \multirow{2}[4]{*}{Road Type} & \multirow{2}[4]{4.5em}{Prediction Horizon} & \multicolumn{2}{C{5em}|}{Public Availability} \\
    \cline{7-8}
    & & & & & & Code  & Data \\
    \hline
    \endfirsthead

    \multicolumn{8}{c}%
    {{\tablename\ \thetable{} -- continued from previous page}} \\

    \hline
    \multirow{2}[4]{*}{Ref} & \multirow{2}[4]{*}{Technique} & \multirow{2}[4]{4.5em}{Prediction Parameter}  & \multirow{2}[4]{*}{Dataset} & \multirow{2}[4]{*}{Road Type} & \multirow{2}[4]{4.5em}{Prediction Horizon} & \multicolumn{2}{C{5em}|}{Public Availability} \\
    \cline{7-8}
    & & & & & & Code  & Data \\
    \hline
    \endhead

    \hline \multicolumn{8}{|r|}{{Continued on next page}} \\ \hline
    \endfoot

    \hline
    \endlastfoot

    \cite{s19102229} & Stacked Auto-encoders & Congestion & Seattle Area Traffic Congestion Status & Freeways & Short & $\times$ & $\checkmark$ \\
    \hline
    \cite{s17040818} & CNN & Speed & Ring roads of Beijing & Freeway & Short & $\times$ & $\times$ \\
    \hline
    \cite{Dai2019}  & CNN   & Flow  & PeMS-D4  & Freeway & Short & $\times$     & $\checkmark$ \\
    \hline
    \cite{LI20191}  & DBN   & Flow  & Highway I-43, Wisconsin  & Freeway & Long & $\times$ & $\times$ \\ \hline
    \cite{Cui2018} & LSTM & Speed & DRIVE Net-Seattle \& INRIX-Seattle & Urban \& Freeway & Short & $\times$     & $\checkmark$ \\
    \hline
    \cite{Chen2021}   & LSTM & Flow  & PeMS-San Diego & Freeway & Short & $\times$     & $\checkmark$ \\ \hline
    \cite{Learning2018} & Spatial-Temporal GCN & Speed & PeMS-D7 & Freeway & Short & $\times$     & $\checkmark$ \\
    \hline
    \cite{Zhao2020}   & Temporal-GCN & Speed & SZ-taxi \& PeMS-Los Angeles & Freeway & Short & $\checkmark$     & $\checkmark$ \\
    \hline
    \cite{Li2021a}   & GCN & Flow  & PEMS-San Fransisco \& San Bernardino & Freeway & Short & $\times$     & $\checkmark$ \\
    \hline
    \cite{Zheng2020}   & Graph multi-attention network (GMAN) & Volume \& Speed & Xiamen \& PeMS-Bay & Urban \& Freeway & Short & $\checkmark$     & $\checkmark$ \\
    \hline
    \cite{Yu2021a}   & Generative adversarial Graph Attention Network & Speed & NavInfo Traffic Data & Urban & Short & $\times$     & $\times$ \\
    \hline
    \cite{Yin2021}   & Multi-stage  attention spatial-temporal graph network & Flow \& Speed & PEMS- San Fransisco \& San Bernardino & Freeway & Short & $\times$     & $\checkmark$ \\
    \hline
    \cite{DeMedrano2020}  & CNN + Auto-encoder & Flow  & Traffic \& weather data from Municipality of Madrid & Urban & Long  & $\checkmark$     & $\checkmark$ \\
    \hline
    \cite{Lu2021}   & ARIMA + LSTM & Flow  & British Gov. & Freeway & Short & $\times$     & $\checkmark$ \\
    \hline
    \cite{Du2020}   & CNN-GRU with attention & Flow  & Highways England & Freeway & Short & $\times$     & $\checkmark$ \\
    \hline
    \cite{Peng2020}  & GA + NN & Volume & PeMS  & Freeway & Short & $\times$     & $\checkmark$ \\
    \hline

    \cite{Li2017} & Diffusion Convolutional RNN & Speed & PeMS-Bay & Freeway & Short & $\checkmark$ & $\checkmark$ \\
    \hline
    \cite{Zheng2020b}  & Gradient–boosted regression tree & Flow  & ERI data in Chongqing & Urban & Short & $\times$     & $\times$ \\
    \hline
    \cite{Cui2020}  & Gated graph wavelet RNN & Speed & INRIX-Seattle & Urban \& Freeway & Short & $\times$     & $\checkmark$ \\
    \hline

\end{longtable}
\end{center}

\normalsize
To fill the research gaps discussed above, a deep learning based approach has been developed to accurately predict traffic speed at multiple time steps. As the proposed approach is a deep learning based approach, so it does not have the short-comings of statistical and machine learning approaches. For instance, statistical techniques assume stationary data and hence can not work on dynamically changing data whereas the proposed approach can easily handle the dynamic variations. Similarly, opposed to machine learning techniques the proposed approach does not require the manual selection of features. It can extract its own features from the raw data. As discussed previously, the main problem with existing deep learning based approaches is that most of them do not consider either spatial or temporal dependencies properly. To solve this problem, in the proposed approach an algorithm has been developed which at a particular instant selects the neighboring sensors from neighboring roads based on current and past traffic similarity and sensor distance from the target sensor. So, the selection of neighboring sensors changes with time and location of the target sensor. After selecting the neighboring sensors for a target sensor, the historical traffic data from these sensors only is used as input to the deep learning model. This helps the deep learning model to better understand the spatio-temporal dependency. Also, the proposed approach uses the concept of latent space mapping by cross-connecting two pre-trained auto-encoders which helps in better feature extraction of source and destination domain. The above reasons combined with optimal choice of historical traffic data prove to be beneficial, hence motivating better performance of proposed approach than existing techniques. The major contributions of the current work are as follows:

\begin{enumerate}[i)]
    \item A deep learning based approach has been developed for multistep traffic speed prediction.
    \item An algorithm has been developed for neighboring sensor selection to take into account the spatio-temporal dependencies of traffic.
    \item The proposed approach uses two cross-connected auto-encoders using latent space mapping which helps in better understanding of source and target domain.
    \item The proposed approach has been tested on real-world traffic data for validating its effectiveness.
\end{enumerate}

The remaining paper has been organized into sections as follows. The proposed methodology has been described in detail in Section~\ref{Sec:Methodology}. Section~\ref{Sec:ExperimentalSetup} provides the description of the real-world dataset and the initial hyper-parameters used for training the proposed approach. The proposed approach is compared with a number of benchmark techniques and the comparison results are discussed in Section~\ref{Sec:Results}. Finally, Section~\ref{Sec:Conclusion} provides the conclusion of the paper along with some future directions.

\section{Methodology}\label{Sec:Methodology}
\begin{figure}[h!]
      \centering
      \includegraphics[width=\linewidth]{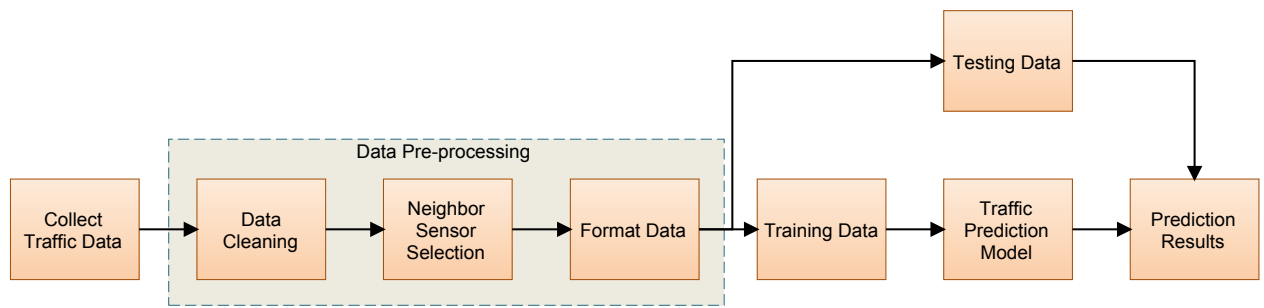}
      \caption{Framework of the proposed methodology}\label{Fig:Methodology}
\end{figure}

The problem of multistep traffic prediction (i.e. traffic prediction at multiple time steps in the future) has been addressed by developing a deep learning approach using the concept of latent space mapping. Figure \ref{Fig:Methodology}, shows the framework of the proposed methodology. Firstly, raw traffic speed data was collected from the loop detector sensors installed on city roads. These loop detectors can provide various traffic-related parameters like traffic speed, volume, and flow, etc. In this work, historical traffic speed has been used to predict future traffic speed. The data was pre-processed to make it suitable for use in training and testing the proposed deep learning model. Certain data cleaning algorithms were employed for removing the various data anomalies, which occur due to the physical limitations of loop detector sensors. After data cleaning, the neighboring sensors of all sensors were selected based on their different attributes (like correlation, and distance, etc.). Then, historical traffic data from all these neighboring sensors was formatted in a particular format to give input to the proposed model. In the proposed model, two auto-encoders were trained separately: one for learning the pattern of historical traffic speed and the other for future traffic speed. After the individual training of both the auto-encoders, the pre-trained encoder of the first auto-encoder and pre-trained decoder of the second auto-encoder was concatenated by adding a latent feature mapping layer in between them. Then, the concatenated model was optimized by training the latent feature mapping layer. Finally, the trained model was used to predict the future traffic speed. The following subsections discuss each part of the methodology in detail.

\subsection{Data Pre-processing}\label{SubSec:DataPreprocessing}
Pre-processing of the collected traffic speed data was required before using the data for training or testing the proposed deep learning model. It was done in three steps as discussed below:

\subsubsection{Data Cleaning}\label{SubSubSec:DataCleaning}
Traffic speed data collected from various loop detectors sensors contain a number of anomalies like missing values due to failure of sensors and presence of outliers due to noise etc. For this, data was cleaned using data cleaning techniques before use. For instance, the loop detector sensors, for which the missing values were more than a particular threshold percentage, were removed from the dataset and for other loop detector sensors, the missing values were replaced with values obtained by linearly interpolating the data. Similarly, the outliers (mainly point outliers which can not be valid speed values for that highway) in the dataset were replaced using a sliding window averaging technique.

\subsubsection{Selection of Neighboring Sensors}\label{SubSubSec:NeighborSelection}
Traffic on a particular road depends on the traffic status of neighboring roads. This is known as the spatial dependency of the traffic. The spatial dependency also changes with time, for instance, it might be possible that at a particular time more traffic is coming from one neighboring road and at another time from other neighboring roads. So, it is very important to cluster the traffic data according to the spatial and temporal dependency. For this, a neighbor sensor selection algorithm (see Algorithm~\ref{Algo:NeighborSelection}) has been used. The algorithm takes multiple input parameters, namely (i) list of all $n$ sensors $\mathds{S}$, (ii) distance matrix $\mathds{D}$ of size $n \times n$, where each element $d_{i,j} \in \mathds{D}$ represents the distance from sensor $i$ to sensor $j$, (iii) traffic speed matrix of the sensors $\mathds{T}_\mathds{S}$ of size $\tau \times n$, where $\tau$ represents the number of time instances for which speed data was recorded and $n$ is the number of sensors, (iv) sensor $p$ for which neighboring sensors are to be identified and (v) time instant $t_0$. Firstly, all the sensors which are less than $\delta$ distance apart from sensor $p$ are selected and stored in $S$. Also, the distance of these sensors from $p$ is stored in $D_S$. Then, traffic speed data of all the sensors $S$ for the time interval of $t_0-\Delta t$ is obtained in $T_S$. Then, using the traffic speed data $T_S$, traffic similarity is calculated between $p$ and each sensor $q \in S$. For traffic similarity, Pearson's correlation coefficient (\cite{Lin1989}) and the absolute mean traffic speed difference were used. So, for each sensor $q \in S$ three attributes were obtained i.e. distance between the sensors $D_S$, Pearson's correlation coefficient $C_S$ and absolute mean traffic speed difference $\triangle_S$. The sensors in $S$ are then ranked using a multi-criteria decision-making technique named Technique for Order of Preference by Similarity to Ideal Solution (TOPSIS) (\cite{Behzadian2012}), based on these three attributes with equal weightage $\lambda$ but with different decision-making criteria $\xi$ i.e. $+1$ for correlation because it should be maximum and $-1$ for distance and absolute mean difference because these two should be minimum. Then, top $m$ sensors were selected as neighboring sensors $\mathds{S}^*$ and returned by the algorithm.

\newlength{\commentWidth}
\setlength{\commentWidth}{7.3cm}
\newcommand{\atcp}[1]{\tcp*[r]{\makebox[\commentWidth]{#1\hfill}}}

\begin{algorithm}[h]
\SetAlgoLined
\DontPrintSemicolon
\KwIn{$\mathds{T}_\mathds{S}$, $\mathds{S}$, $\mathds{D}$, $p$, $t_0$}
\KwOut{$\mathds{S}^*$}

    \SetKwFunction{FMain}{NeighborSensorSelection}
    \SetKwProg{Fn}{Function}{:}{}
    \Fn{\FMain{$\mathds{T}_\mathds{S}$, $\mathds{S}$, $\mathds{D}$, $p$, $t_0$}}
    {
        initialize distance threshold $\delta$\;
        $S \longleftarrow \mathds{S}[\mathds{D}[p,:] < \delta]$         \atcp{Get sensors with distance $< \delta$}
                                                              \atcp{from $p$}
        $D_S \longleftarrow \mathds{D}[p,S]$                  \atcp{Get distance of sensors $S$}
                                                              \atcp{from $p$}
        $T_S \longleftarrow \mathds{T}_\mathds{S}[t_0-\Delta t:t_0,S]$   \atcp{Get past traffic data of}
                                                              \atcp{sensors $S$ for $\Delta t$ duration}
        \ForEach{$q \in S$}
        {
            $C_S[q]\longleftarrow corr(T_S[:,p],T_S[:,q])$               \atcp{Find traffic correlation}
            $\triangle_S[q] \longleftarrow \big| \overline{T_S[:,p]} - \overline{T_S[:,q]} \big|$   \atcp{Find absolute mean difference}
        }
        $\alpha_S \longleftarrow concatenate(C_S,D_S,\triangle_S)$   \atcp{Concatenate the attributes}
        $\xi \longleftarrow [+1,-1,-1]$                              \atcp{Attributes to be maximized(+)}
                                                                     \atcp{or minimized(-)}
        $\lambda \longleftarrow [1,1,1]$                             \atcp{Weightage of the attributes}
        $R_S \longleftarrow topsis(\alpha_S,\lambda,\xi)$            \atcp{Rank the sensors using Topsis}
        $\mathds{S}^* \longleftarrow R_S[1:m]$                    \atcp{Select top $m$ sensors}
        \textbf{return} $\mathds{S}^*$\;
    }
\textbf{end}
\caption{Algorithm for selecting the neighboring sensors}
\label{Algo:NeighborSelection}
\end{algorithm}

\subsubsection{Formatting the traffic data}\label{SubSubSec:FormatData}
After selecting the neighboring sensors for a particular sensor $p$ at a particular time instance $t_0$, past traffic data for all these neighboring sensors $\mathds{S}^*$ was formatted in a particular format before using it for training and testing the proposed deep learning model. The approach discussed by \cite{s17040818} motivated the authors to convert historical traffic data into images hence, generating a spatio-temporal representation. The idea here is to capture historical traffic pattern into the traffic data matrix (say, $X$ of size $l \times m \times c$), created from the past traffic speed data of all selected neighboring sensors. Here, $l$ represents the number of time instances in the time interval $\Delta t$, $m$ represents the number of selected nearby sensors and $c$ is the number of channels. In this work, $c$ was taken as 4 and each channel corresponds to the traffic data of a different day, as shown in Figure \ref{Fig:TrafficDataMatrix}. The first channel has data of the current day (say $d$), the second channel contains data for the past day i.e. $d-1$ and the third and fourth channel contains data for the same day of the week as the current day but of past two weeks i.e. $d-7$ and $d-14$. This was done to capture the weekly traffic pattern. The time instances for which traffic data was used for each channel are also different i.e. first channel contains data for time instances between $t_0-\Delta t$ to $t_0$ whereas other channels contain data for time instances between $t_0-\frac{\Delta t}{2}$ to $t_0+\frac{\Delta t}{2}$. This helps the deep learning model in understanding that how was the traffic speed pattern after time instant $t_0$ in the past day and past two weeks. The traffic speed data matrix $X$ was used as input for the proposed traffic prediction model. The output of the traffic prediction model is future traffic speed vector $Y = [y_1, y_2, \cdots y_n]^T$ of sensor $p$. Here, $y_i$ represents the traffic speed at sensor location $p$ at time instance $t_0 + i$. The traffic data matrix $X$ and $Y$ were normalized into the range of [0,1] using Eq. \eqref{Eq:Normalize}.

\begin{equation}\label{Eq:Normalize}
    \hat{z^j} = \frac{z^j - \min(z)}{\max(z) - \min(z)}
\end{equation}

In above equation, $\hat{z^j}$, $z^j$, $\min(z)$ and $\max(z)$ represent the normalized traffic data of $j^{th}$ sensor, traffic data of $j^{th}$ sensor, minimum and maximum of traffic data of all sensors, respectively. The normalized traffic speed data pair $(X,Y)$ was then used for training and testing the proposed deep learning model.

\begin{figure}[h!]
  \centering
  \begin{subfigure}[t]{.48\linewidth}
      \centering
      \begin{math}
          \begin{bmatrix}
            x_{t_0-\Delta t,1}^d     & x_{t_0-\Delta t,2}^d     & \cdots & x_{t_0-\Delta t,m}^d \\
            x_{t_0-\Delta t+1,1}^d   & x_{t_0-\Delta t+1,2}^d   & \cdots & x_{t_0-\Delta t+1,m}^d \\
            \vdots                   & \vdots                   & \ddots & \vdots  \\
            x_{t_0,1}^d              & x_{t_0,2}^d              & \cdots & x_{t_0,m}^d
          \end{bmatrix}
      \end{math}
      \caption{$X(:,:,1)$}
  \end{subfigure}
  \hfill
  \begin{subfigure}[t]{.48\linewidth}
      \centering
      \begin{math}
          \begin{bmatrix}
            x_{t_0-\frac{\Delta t}{2},1}^{d-1}     & x_{t_0-\frac{\Delta t}{2},2}^{d-1}     & \cdots & x_{t_0-\frac{\Delta t}{2},m}^{d-1} \\
            x_{t_0-\frac{\Delta t}{2}+1,1}^{d-1}   & x_{t_0-\frac{\Delta t}{2}+1,2}^{d-1}   & \cdots & x_{t_0-\frac{\Delta t}{2}+1,m}^{d-1} \\
            \vdots                         & \vdots                         & \ddots & \vdots  \\
            x_{t_0+\frac{\Delta t}{2},1}^{d-1}     & x_{t_0+\frac{\Delta t}{2},2}^{d-1}     & \cdots & x_{t_0+\frac{\Delta t}{2},m}^{d-1}
          \end{bmatrix}
      \end{math}
      \caption{$X(:,:,2)$}
  \end{subfigure}
  \begin{subfigure}[t]{.48\linewidth}
      \centering
      \begin{math}
          \begin{bmatrix}
            x_{t_0-\frac{\Delta t}{2},1}^{d-7}     & x_{t_0-\frac{\Delta t}{2},2}^{d-7}     & \cdots & x_{t_0-\frac{\Delta t}{2},m}^{d-7} \\
            x_{t_0-\frac{\Delta t}{2}+1,1}^{d-7}   & x_{t_0-\frac{\Delta t}{2}+1,2}^{d-7}   & \cdots & x_{t_0-\frac{\Delta t}{2}+1,m}^{d-7} \\
            \vdots                         & \vdots                         & \ddots & \vdots  \\
            x_{t_0+\frac{\Delta t}{2},1}^{d-7}     & x_{t_0+\frac{\Delta t}{2},2}^{d-7}     & \cdots & x_{t_0+\frac{\Delta t}{2},m}^{d-7}
        \end{bmatrix}
      \end{math}
      \caption{$X(:,:,3)$}
  \end{subfigure}
  \hfill
  \begin{subfigure}[t]{.48\linewidth}
      \centering
      \begin{math}
          \begin{bmatrix}
            x_{t_0-\frac{\Delta t}{2},1}^{d-14}     & x_{t_0-\frac{\Delta t}{2},2}^{d-14}     & \cdots & x_{t_0-\frac{\Delta t}{2},m}^{d-14} \\
            x_{t_0-\frac{\Delta t}{2}+1,1}^{d-14}   & x_{t_0-\frac{\Delta t}{2}+1,2}^{d-14}   & \cdots & x_{t_0-\frac{\Delta t}{2}+1,m}^{d-14} \\
            \vdots                          & \vdots                          & \ddots & \vdots  \\
            x_{t_0+\frac{\Delta t}{2},1}^{d-14}     & x_{t_0+\frac{\Delta t}{2},2}^{d-14}     & \cdots & x_{t_0+\frac{\Delta t}{2},m}^{d-14}
          \end{bmatrix}
      \end{math}
      \caption{$X(:,:,4)$}
  \end{subfigure}

  \caption{The channels of formatted traffic data matrix $X$}\label{Fig:TrafficDataMatrix}
\end{figure}

\subsection{Traffic Prediction Model}\label{SubSec:TrafficPredictionModel}
\begin{figure}[H]
      \centering
      \includegraphics[scale=0.42]{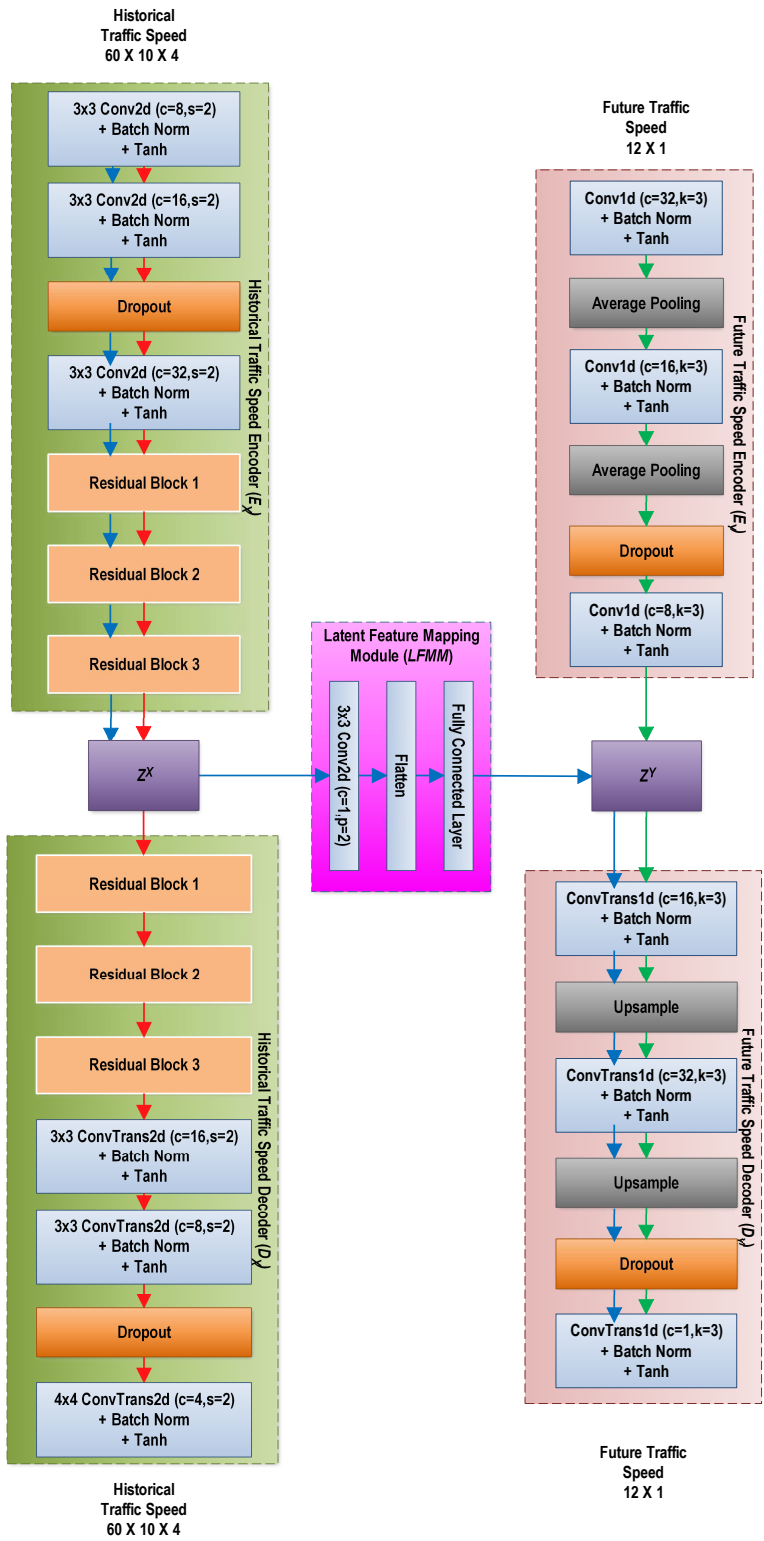}
      \caption{Architecture of Traffic Prediction Model}\label{Fig:ArchitectureDiagram}
\end{figure}

A deep learning based model inspired by Deep Auto-Encoders (DAEs) has been developed for multistep traffic speed prediction. An auto-encoder has two main components, namely, an encoder and a decoder. The encoder is used to learn the internal representation of the input. This internal representation is also known as the latent space representation, which can also be used as a feature vector. The decoder does the reverse process and learns to convert the internal representation back to the input. Multiple researchers have used DAEs for robust feature extraction and using the features for classification or regression tasks. \cite{Vincent2008} used auto-encoders to extract robust features by giving the corrupted inputs. Similarly, \cite{Masci2011} used the hierarchical features extracted using auto-encoders to initialize the convolutional neural network. Although they have used auto-encoders for different problems, it has shown the effectiveness of auto-encoders for extracting robust features. So, in this work pre-trained auto-encoders are used to predict the multistep traffic speed. Figure~\ref{Fig:ArchitectureDiagram}, shows the architecture of the proposed deep learning model. Two different DAEs were used in the proposed architecture: one for historical traffic speed and the other for future traffic speed. The encoder of the first DAE, $E_X$, converts the historical traffic speed $X$ into the corresponding latent representation $Z^{X}$, i.e. $E_{X}: X \rightarrow Z^{X}$. On the contrary, the decoder $D_X$ recreate the historical traffic speed from its latent representation, i.e. $D_{X}: Z^{X} \rightarrow X$. Similarly, the encoder and decoder of other DAE converts future traffic speed $Y$ to its corresponding latent representation $Z^Y$ and then back to the future traffic speed. The extraction of $i^{th}$ latent feature map ($l^i$) for input $I$ can be expressed as:
\begin{equation}
l^i = \sigma(I * W^i + b^i)
\end{equation}
where $\sigma$, $W^i$, $b^i$ and $*$ represent the activation function, weight, bias, and convolution operation. Similarly, the reconstruction from latent representation is obtained using:
\begin{equation}
\hat{I} = \sigma\Bigg(\int_{i \in L} l^i * \hat{W} + c\Bigg)
\end{equation}
where $L$, $\hat{W}$, and $c$ represent the list of latent feature maps, transpose convolution operation, and bias for the input channel. So, the two DAEs try to extract the features of individual domains in terms of their corresponding latent representation. This is done by minimizing the mean square error loss between the input $I$ and reconstructed output $\hat{I}$ of both the DAEs which can be expressed as follows:
\begin{equation}
\mathcal{L}_{DAE} : f(I,\hat{I}) = ||I-\hat{I}||^2
\end{equation}

The idea was to use both the pre-trained DAEs, i.e. $DAE_X: E_X - D_X$ and $DAE_Y: E_Y - D_Y$, by cross-connecting the encoder $E_X$ with decoder $D_Y$ for predicting the future traffic speed from historical traffic speed. The cross-connection was done using Latent Feature Mapping Module $LFMM$, which maps the latent representation of historical traffic speed $Z^X$ to the latent representation of future traffic speed $Z^Y$ i.e. $FM: Z^X \rightarrow Z^Y$. As both the DAEs were trained separately, the two latent space representations i.e. $Z^X$ and $Z^Y$ do not lie in the same dimensional space. So, the $LFMM$ fills the gap of properly mapping the two latent spaces. Since the encoder $E_X$ was already trained to convert the historical traffic speed $X$ into corresponding latent representation $Z^X$, it learns to effectively extract the critical features of $X$ in the traffic prediction learning phase. Similarly, the decoder $D_Y$ also learns to regenerate future speed $Y$ from latent representation $Z^Y$ while preserving the normal characteristics of $Y$ due to the pre-training effect. In Figure~\ref{Fig:ArchitectureDiagram}, the flow of data for both the DAEs, i.e. $DAE_X$ and $DAE_Y$ is shown by red and green arrows, respectively. Similarly, the flow of data of cross-connected DAEs is represented using blue arrows. The following subsections discuss the architecture of both the DAEs and $LFMM$ in detail.

\subsubsection{Deep Auto-Encoders (DAEs)}
As discussed earlier, there are two DAEs ($DAE_X$ and $DAE_Y$) in the proposed architecture. The encoder of $DAE_X$ has three convolutional layers followed by batch normalization and $Tanh$ activation layer. The three convolutional layers use convolutional kernels of size $3 \times 3$ with a stride of 2 and padding of 1. Past research in deep learning has proved that the deeper networks can achieve better performance but simply stacking the layers makes the training process difficult due to the problem of vanishing gradient. So, to overcome this problem the concept of residual learning using residual blocks was introduced in \cite{He2016}. Inspiring by the success of residual learning, three residual blocks were used in the encoder $E_X$. Each residual block has two convolutional layers with a kernel size of $3 \times 3$, stride of 1, and padding of 1 followed by batch normalization and non-linear activation ReLU. Each residual block has a connection that adds input of residual block to the batch normalized output of the second convolutional layer of the block. To avoid overfitting of the proposed model, a dropout layer was also used. The decoder $D_X$ also has a similar structure but in reverse order. To reverse the convolution effect, the transposed convolutional layers were used instead of convolutional layers.

The auto-encoder $DAE_Y$ has a simple architecture of a convolutional auto-encoder. The encoder $E_Y$ has three convolutional layers each followed by batch normalization and $Tanh$ layer. Each convolutional layer has kernels of size 3, stride, and padding of 1. Two average pooling layers were also used to keep the average effect of features and reduce dimension. The decoder $D_Y$ also has a similar architecture with convolutional layers replaced with transposed convolutional layers and average pooling layers replaced with up-sampling layers. Dropout layers were used both in the encoder and decoder to avoid overfitting.

\subsubsection{Latent Feature Mapping Module ($LFMM$)}
The two DAEs were cross-connected using $LFMM$, which maps the latent features $Z^X$ of historical traffic speed to the latent features $Z^Y$ of future traffic speed. The $LFMM$ has three layers in sequence i.e. a convolution layer followed by a flatten and a fully connected layer. The convolution layer has convolutional kernels of size $3 \times 3$ with 1 stride and padding of 2 rows and 2 columns. As shown in Figure~\ref{Fig:ArchitectureDiagram}, the blue arrows show the flow of data for the cross-connected network. The entire cross-connected network was fine-tuned after pre-training the individual DAEs separately.

\section{Experimental Setup}\label{Sec:ExperimentalSetup}
This section describes the dataset used and hyper-parameters initialization along with other details for training the proposed approach.

\subsection{Datasets}\label{SubSec:DataSets}
\begin{figure}[t]
      \centering
      \begin{subfigure}[]{0.45\textwidth}
        \centering
        \includegraphics[scale = 0.5]{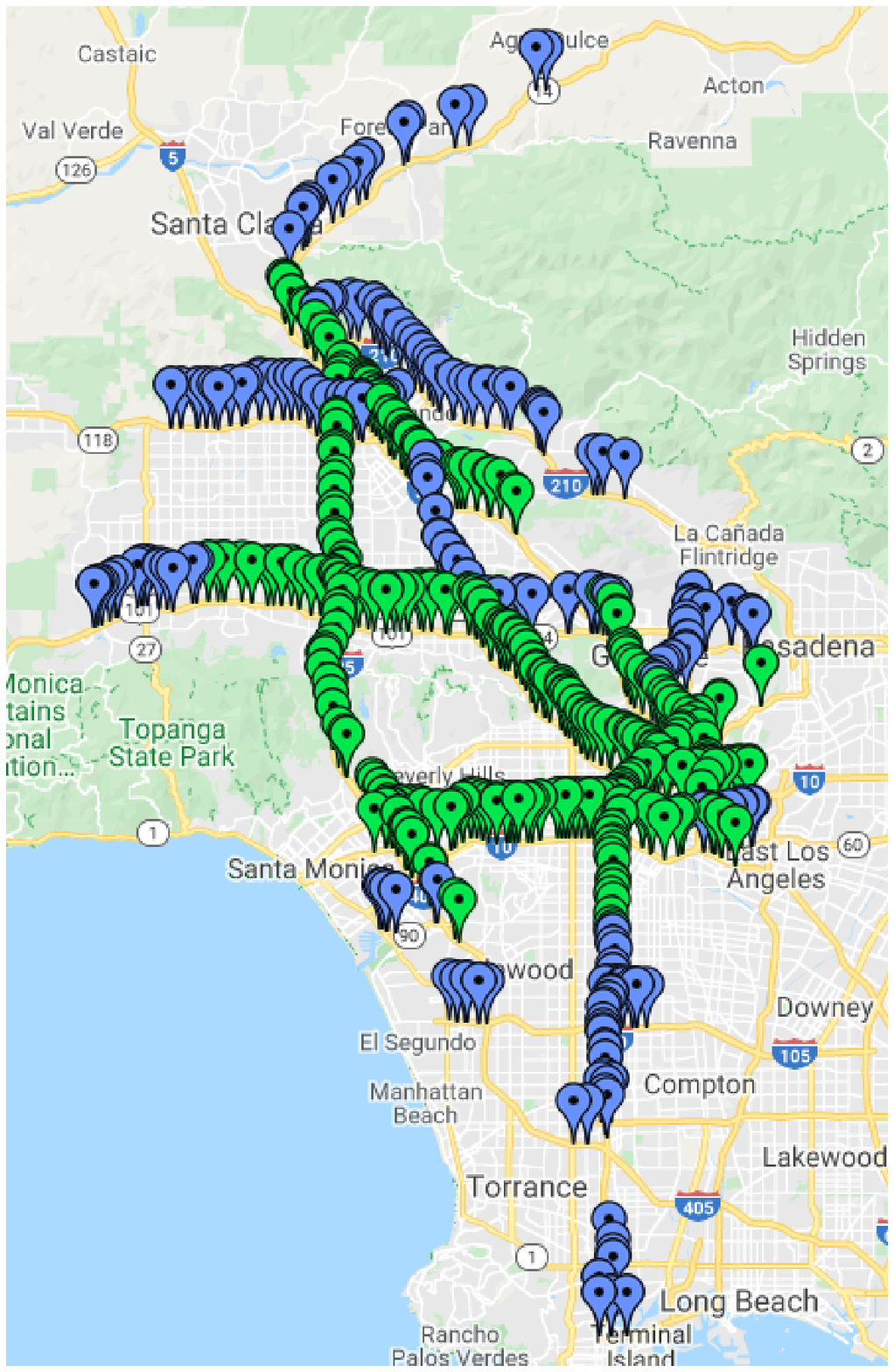}
        \caption{Sensors of Los Angeles}\label{SubFig:SelectedSensorsLosAngeles}
      \end{subfigure}
      \hfill
      \begin{subfigure}[]{0.45\textwidth}
        \centering
        \includegraphics[scale = 0.55]{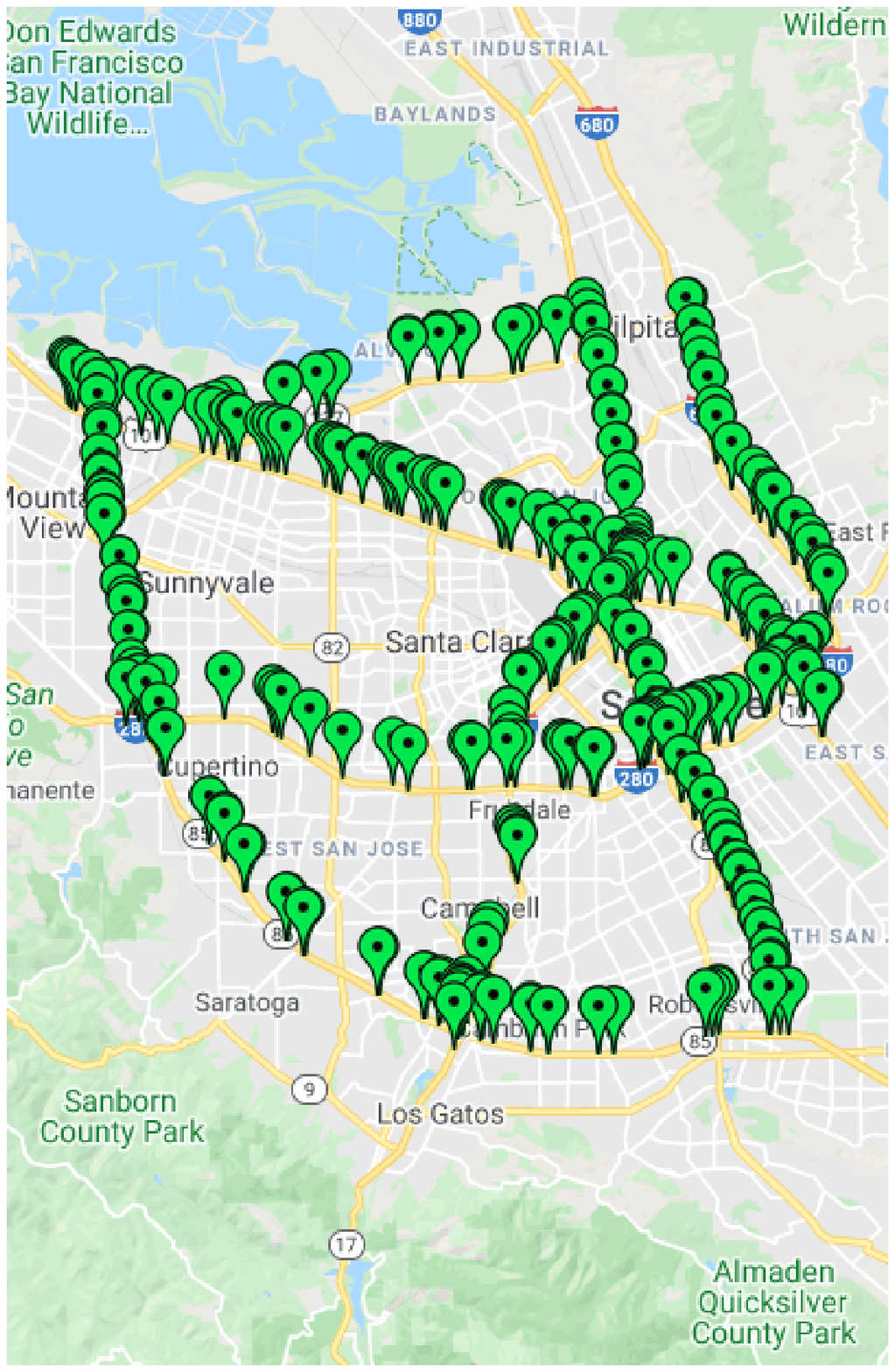}
        \caption{Sensors of Bay Area}\label{SubFig:SelectedSensorsBayArea}
      \end{subfigure}

      \caption{Loop detector sensors on the map of Los Angeles and Bay Area}\label{Fig:SensorsOnMap}
\end{figure}

Traffic data from a very popular data source, available through the web-portal of the California Department of Transportation Performance Measurement System (PeMS) (\cite{PeMS2020}) is used. The data source provides a large amount of traffic data from the loop detectors installed on the freeways of the state of California. From PeMS, traffic data from two areas namely, Los Angeles and Bay Area, have been taken and used for training and testing the proposed methodology. Figure~\ref{Fig:SensorsOnMap} shows the location of loop detector sensors on the map of Los Angeles and Bay Area.

In Figure~\ref{SubFig:SelectedSensorsLosAngeles}, there are 660 sensors, shown on the map with markers, on the mainline of freeways of Los Angeles. Out of these sensors, 382 sensors, shown with green markers, are selected which cover the five freeways of Los Angeles, namely, I5, I10, I110, I405, and US101. Similarly, Figure~\ref{SubFig:SelectedSensorsBayArea} shows the location of 325 sensors on the map of Bay Area. The proposed traffic prediction model was trained and tested to predict future traffic speed for these selected sensors. As explained in Section~\ref{SubSubSec:NeighborSelection}, the neighboring sensors for all these 382 sensors were obtained using Algorithm~\ref{Algo:NeighborSelection}. The algorithm can select any sensor as the neighboring sensor depending upon the traffic pattern. As there are two different locations for which traffic data has been used, from now onwards the data from Los Angeles and Bay Area will be denoted as $PEMS-Los$ and $PEMS-Bay$, respectively.

For predicting the traffic speed of the next one hour at a particular sensor last five hours of traffic data was used. As traffic speed data at every 5-min interval is available from the data source and top 10 sensors were selected using the neighbor selection algorithm, the input traffic speed matrix $X$ has the dimension of $60 \times 10 \times 4$ and output traffic speed matrix $Y$ has the dimension of $12 \times 1$. $PEMS-Los$ contains traffic data for the period of two months from 1st June, 2017 to 31st July, 2017 and $PEMS-Bay$ has traffic data for the period of six months from 1st Jan, 2017 to 30th Jun, 2017. 70\% of these datasets was used for training the proposed model and the remaining 30\% was used for testing. As traffic at night is very less and does not require forecasting, so for the experiments in the current study traffic data from 7:00 am to 10:00 pm only has been used.

\subsection{Hyperparameters}\label{SubSec:Hyperparameters}
The proposed model has two DAEs which were trained separately. Then, the two DAEs were cross-connected using $LFMM$ and the fine-tuning was done. For training the DAEs and the cross-connected network, all the fully connected, convolutional, and transposed convolutional layers were initialized with Xavier initialization (\cite{glorot2010understanding}). It initializes the weight of a layer using random numbers from a uniform distribution with the limits of $\left[-\sqrt{\frac{6}{fan_{in}+fan_{out}}}, \sqrt{\frac{6}{fan_{in}+fan_{out}}}\right]$. Here, the variables $fan_{in}$ and $fan_{out}$ represent the count of input connections with the layer and count of output connections from the layer, respectively. Adam optimization algorithm (\cite{kingma2014adam}) was used for finding the optimized solution with an initial learning rate of 0.001 and batch size of 128. The adam's algorithm at each step calculates the exponential running average of gradients and square of gradients. To control the decay of these running averages the parameters $\beta_1$ and $\beta_2$ were initialized to 0.9 and 0.999, respectively. During training to avoid by zero epsilon $\epsilon$ was set to a small value of $10^{-8}$. Mean Square Error (MSE) was used as a loss function during training. There are some dropout layers also in the proposed traffic prediction model which were used as regularizers to avoid overfitting. For all the dropout layers the dropout probability was set 0.2, as the experiments show that increasing the drop rate further has a negative effect on the model's performance.

\section{Results and Discussion}\label{Sec:Results}
In this section, the results obtained from experiments have been discussed. First, the traffic speed data from $PEMS-Los$ is used to show the effectiveness of the proposed approach, by comparing the results with several state-of-the-art techniques which include machine learning (ANN, kNN, and XGBoost) and deep learning techniques (LSTM and CNN). These techniques are implemented for comparison as discussed below:

\begin{enumerate}[i)]
    \item \emph{ANN:} A three-layered Artificial Neural Network (ANN) (\cite{Sharma2018}) was developed for traffic speed prediction. Each neuron in ANN was activated using the sigmoid function. As ANN does not differentiate the input variables across time, it was not able to capture the temporal dependencies.
    \item \emph{kNN:} A k-Nearest Neighbor (kNN) (\cite{Cai2016}) approach finds the top $k$ similar observations from the training set based on the Euclidean distance. Then, the future traffic speed prediction is calculated using the weighted sum of the corresponding future traffic speed of $k$ selected observations. The hyper-parameter $k$ is selected using cross-validation by varying the value of $k$ from 5 to 20.
    \item \emph{XGBoost:} XGBoost (\cite{10.1145/2939672.2939785}) is a quite popular machine learning technique that has already been applied to different tasks with outstanding performance. To implement XGBoost, all the input features were reshaped to a vector and then used as the input for training.
    \item \emph{CNN:} Convolutional Neural Network (CNN) (\cite{s17040818}) has been applied to predict traffic speed by reshaping the traffic data into matrices. Although CNNs are capable of learning complex patterns from the data, they can not take temporal information into account.
    \item \emph{LSTM:} Long Short Term Memory (LSTM) (\cite{GU20191}) is a very popular deep learning method and has widely been used for time series forecasting. LSTM is capable of taking temporal dependencies into account but can not capture the spatial information.
\end{enumerate}

The most appropriate inputs were selected for the aforementioned techniques to ensure a fair comparison. The traffic data of the last 5 hours (i.e. past 60 time-steps) of the target sensor was reshaped to form a vector and given as input to ANN, kNN, XGBoost, and LSTM which provide a prediction of the next 1 hour (i.e. 12 time-steps) as output. All these models were trained with a number of such inputs taken at different time instances for all the 382 sensors of $PEMS-Los$. The best hyper-parameters for different models were chosen using cross-validation, for instance, the value of $k=17$ for kNN gave the best performance. For training the CNN, the same matrices were used, which were used for training the proposed approach. The CNN model contains three ReLU activated convolutional layers each followed by an average pooling layer and then a dense layer has been used to take the output. The input traffic data was normalized into the range of [0,1] before using for training the models.

The performance of the above said state-of-the-art techniques are compared with the proposed approach using three standard evaluation metrics, namely Mean Absolute Error, Root Mean Square Error (RMSE), and Mean Absolute Percentage Error (MAPE). These can be calculated using the following equations:

\begin{equation}
    MAE = \frac{1}{N}\sum_{i=1}^{i=N} \big| x_{act}^i - x_{pred}^i \big|
\end{equation}
\begin{equation}
    RMSE = \sqrt{\frac{1}{N}\sum_{i=1}^{i=N} \big( x_{act}^i - x_{pred}^i \big)^2}
\end{equation}
\begin{equation}
    MAPE = \Bigg(\frac{1}{N}\sum_{i=1}^{i=N} \bigg|\frac{x_{act}^i - x_{pred}^i}{x_{act}^i}\bigg| \Bigg) * 100
\end{equation}

In all of the above equations, $N$ is the total number of samples, $x_{act}^i$ and $x_{pred}^i$ represent the actual and predicted value. The evaluation metric MAE and RMSE can measure the absolute deviation of predicted values from actual values whereas MAPE can measure the relative deviation. Table~\ref{Tab:Comparison} shows the performance comparison results of the proposed approach and other benchmark techniques at five different prediction time horizons for traffic data of $PEMS-Los$. From the table, it can be easily concluded that the proposed approach has the best performance with the least values at all of the prediction time horizons except for the 5-min ahead prediction. Although the performance of the proposed approach is not best for 5-min prediction, it is comparable to the performance of other techniques. It can also be observed from the table that as the prediction horizon increases the performance error for each technique also increases but the increase in error with prediction horizon is quite less for the proposed approach as compared to other benchmark techniques. The main reason for this is the ability of the proposed approach to effectively capture spatial and temporal features of the traffic by selecting neighboring sensors at a particular instant based on traffic similarity and distance.

\begin{table}[H]
    \centering
    \caption{Performance comparison of the proposed approach with other popular approaches using $PEMS-Los$ dataset}
    \label{Tab:Comparison}

    \begin{tabular}{|c|c|c|c|c|c|c|}
        \hline
        Technique               & Metric     & 5-min    & 15-min    & 30-min    & 45-min    & 60-min            \\ \hline

        \multirow{3}{*}{ANN}    & MAE        & 1.51     & 2.78      & 3.93      & 4.80      & 5.42              \\ 
                                & RMSE       & 2.58     & 4.90      & 6.92      & 8.14      & 9.03              \\ 
                                & MAPE       & 3.69     & 7.12      & 10.75     & 13.38     & 15.53             \\ \hline

        \multirow{3}{*}{kNN}    & MAE        & 2.28     & 3.12      & 4.02      & 4.74      & 5.29              \\ 
                                & RMSE       & 3.90     & 5.46      & 7.00      & 8.04      & 8.77              \\ 
                                & MAPE       & 5.99     & 8.33      & 11.08     & 13.41     & 15.16             \\ \hline

        \multirow{3}{*}{XGBoost}& MAE        & \textbf{1.27}     & 2.39      & 3.44      & 4.19      & 4.75     \\ 
                                & RMSE       & \textbf{2.13}     & 4.11      & 5.89      & 6.97      & 7.73     \\ 
                                & MAPE       & \textbf{3.09}     & 6.10      & 9.04      & 11.35     & 13.09    \\ \hline

        \multirow{3}{*}{CNN}    & MAE        & 1.92     & 2.88      & 3.62      & 4.09      & 4.41              \\ 
                                & RMSE       & 2.85     & 4.70      & 6.12      & 6.82      & 7.25              \\ 
                                & MAPE       & 4.77     & 7.47      & 9.88      & 11.37     & 12.37             \\ \hline

        \multirow{3}{*}{LSTM}   & MAE        & 1.52     & 2.62      & 3.43      & 3.96      & 4.55              \\ 
                                & RMSE       & 2.48     & 4.38      & 5.85      & 6.73      & 7.55              \\ 
                                & MAPE       & 3.71     & 6.56      & 8.87      & 10.43     & 12.11             \\ \hline

        \multirow{3}{*}{Proposed Approach}  & MAE   & 1.64     & \textbf{2.26}      & \textbf{2.57}      & \textbf{2.75}      & \textbf{3.09}              \\ 
        & RMSE  & 2.58     & \textbf{3.73}      & \textbf{4.51}      & \textbf{4.88}      & \textbf{5.38}       \\ 
        & MAPE  & 4.43     & \textbf{6.03}      & \textbf{6.81}      & \textbf{7.32}      & \textbf{8.44}       \\ \hline
        \multicolumn{7}{|l|}{\footnotesize{Values in bold represent the best ones}}                             \\ \hline
    \end{tabular}

\end{table}

\begin{figure}[H]
  \centering
  \begin{subfigure}[]{0.32\textwidth}
    \centering
    \includegraphics[width=\textwidth]{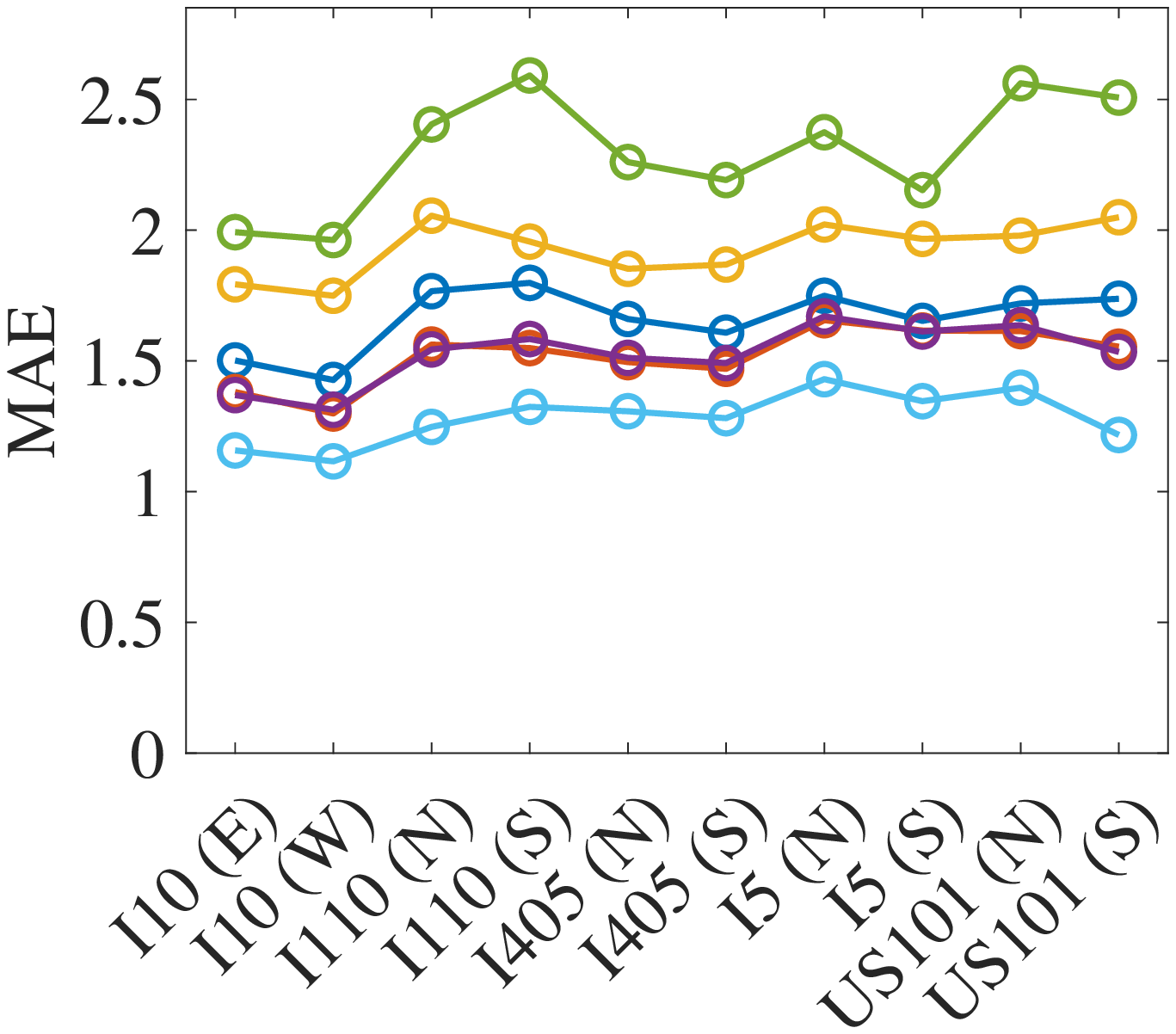}
    \caption{MAE for prediction horizon of 5-min}\label{SubFig:MAE_5}
  \end{subfigure}
  \hfill
  \begin{subfigure}[]{0.32\textwidth}
    \centering
    \includegraphics[width=\linewidth]{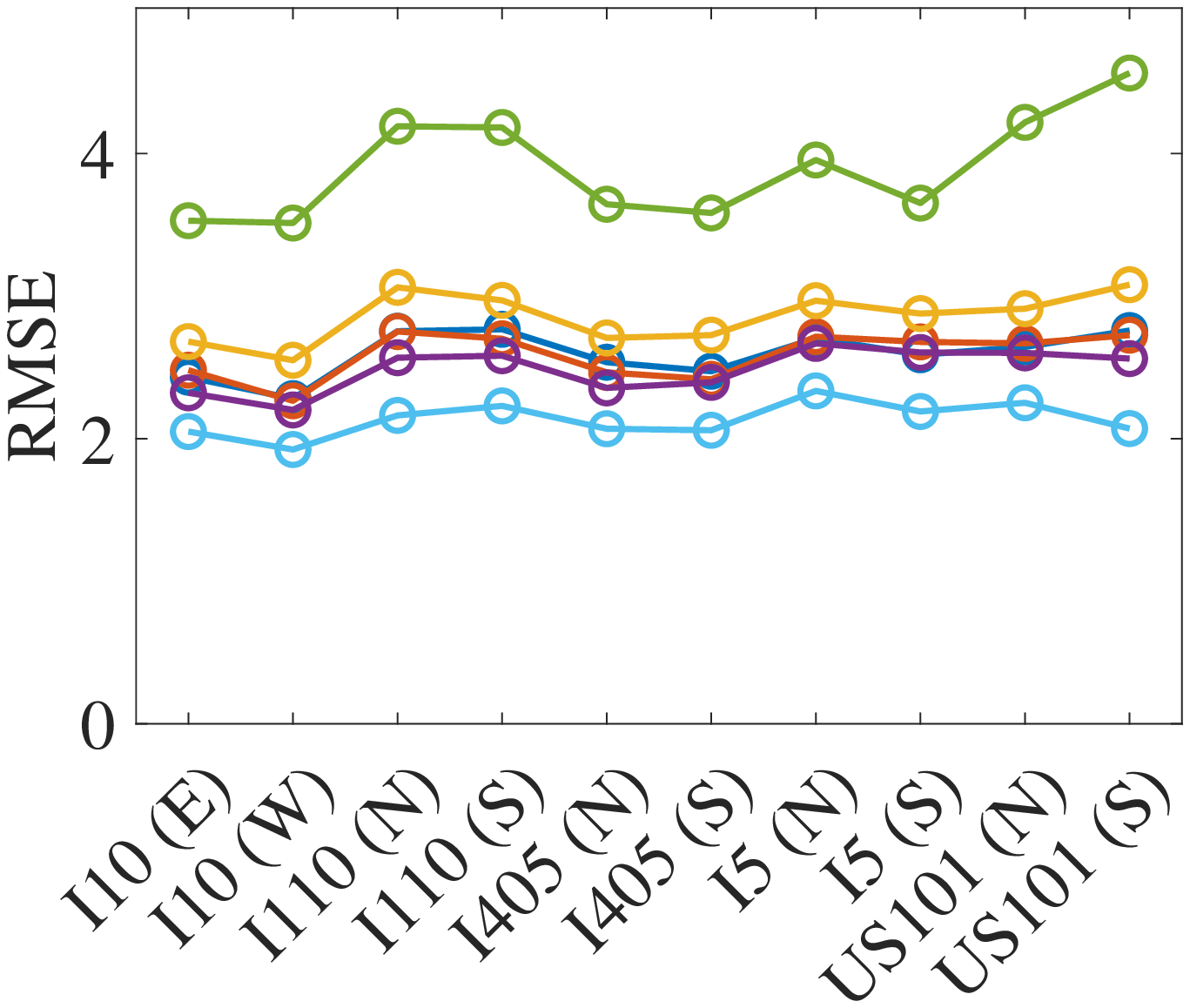}
    \caption{RMSE for prediction horizon of 5-min}\label{SubFig:RMSE_5}
  \end{subfigure}
  \hfill
  \begin{subfigure}[]{0.32\textwidth}
    \centering
    \includegraphics[width=\linewidth]{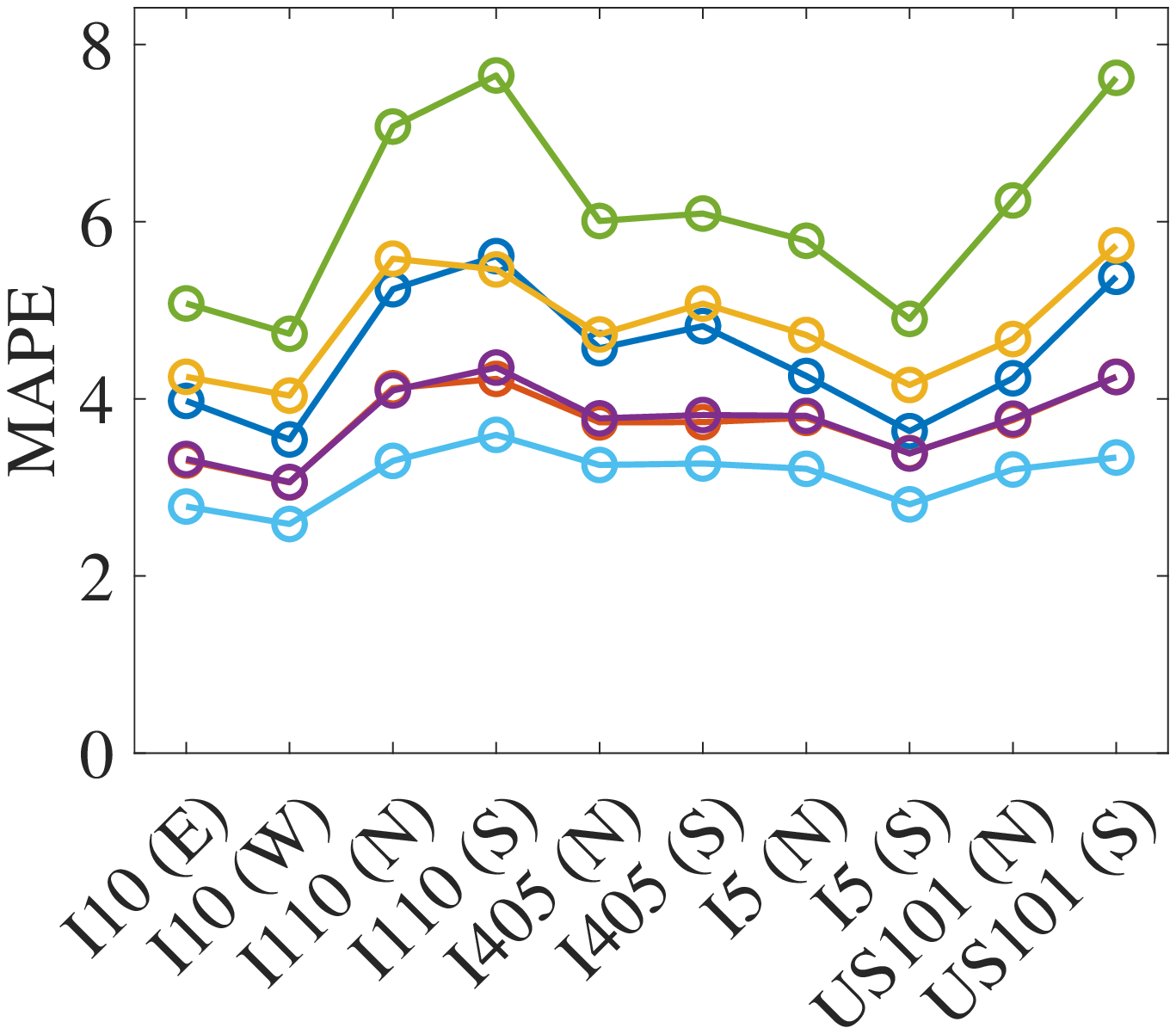}
    \caption{MAPE for prediction horizon of 5-min}\label{SubFig:MAPE_5}
  \end{subfigure}
  \begin{subfigure}[]{0.32\textwidth}
    \centering
    \includegraphics[width=\linewidth]{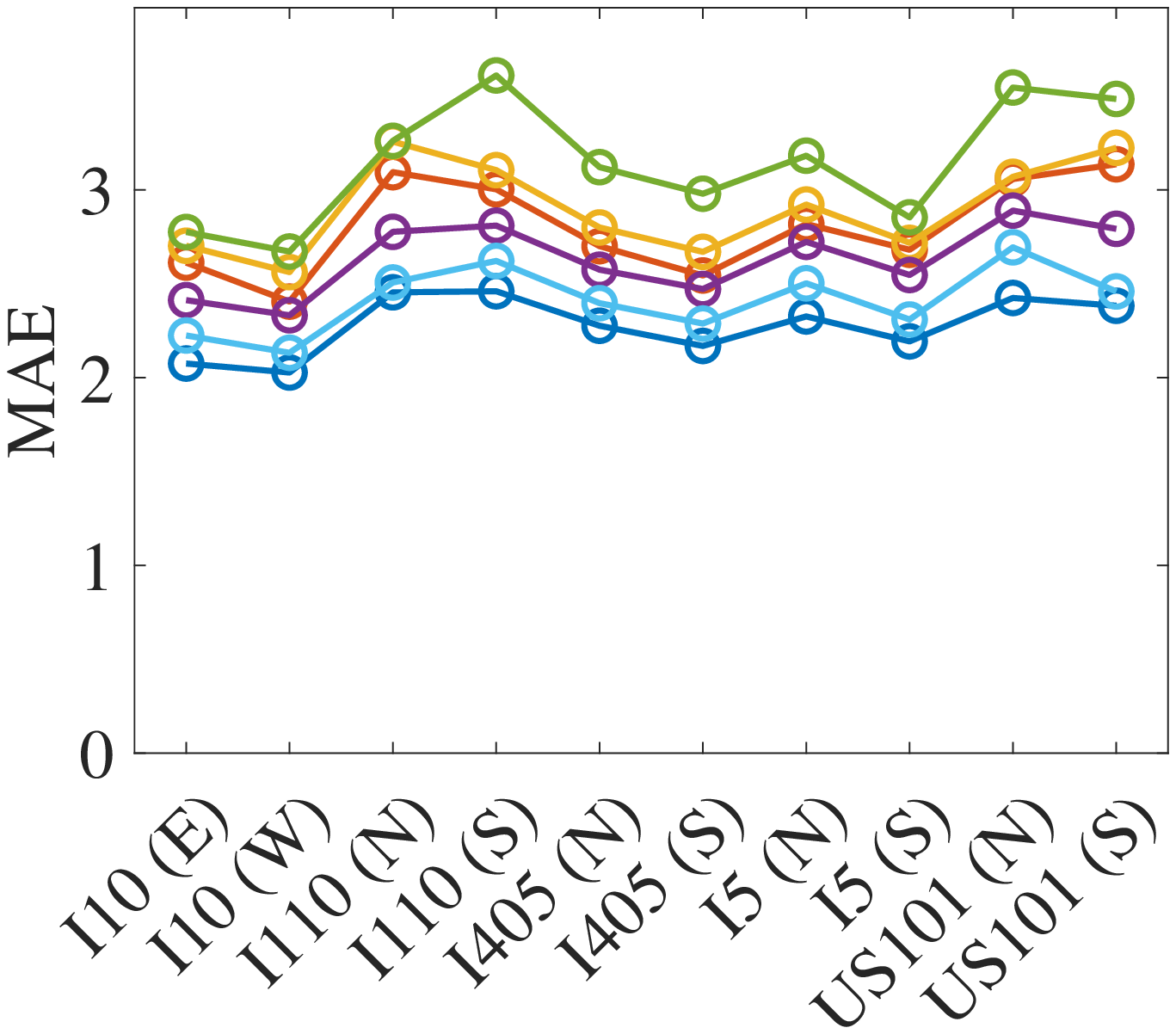}
    \caption{MAE for prediction horizon of 15-min}\label{SubFig:MAE_15}
  \end{subfigure}
  \hfill
  \begin{subfigure}[]{0.32\textwidth}
    \centering
    \includegraphics[width=\linewidth]{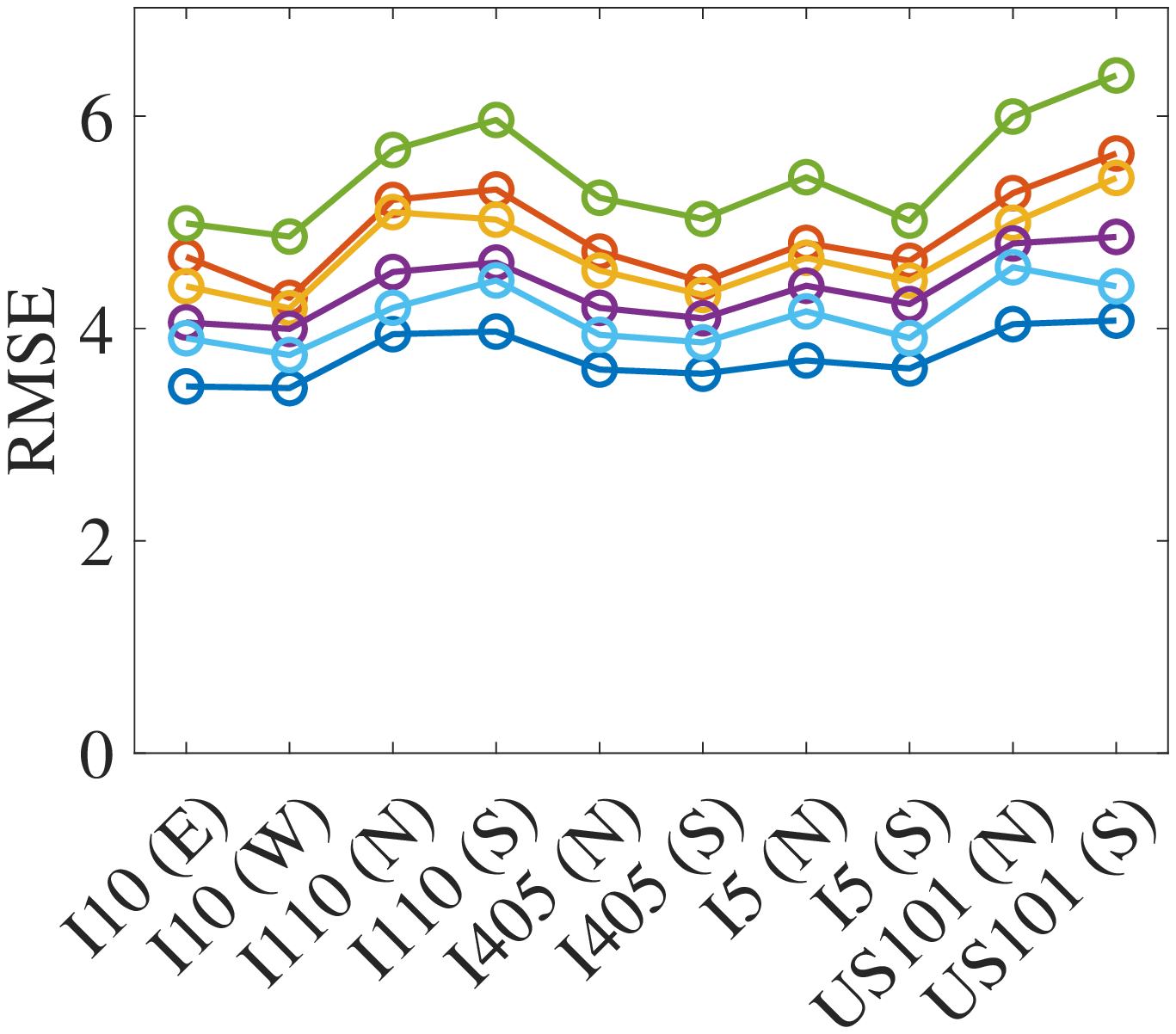}
    \caption{RMSE for prediction horizon of 15-min}\label{SubFig:RMSE_15}
  \end{subfigure}
  \hfill
  \begin{subfigure}[]{0.32\textwidth}
    \centering
    \includegraphics[width=\linewidth]{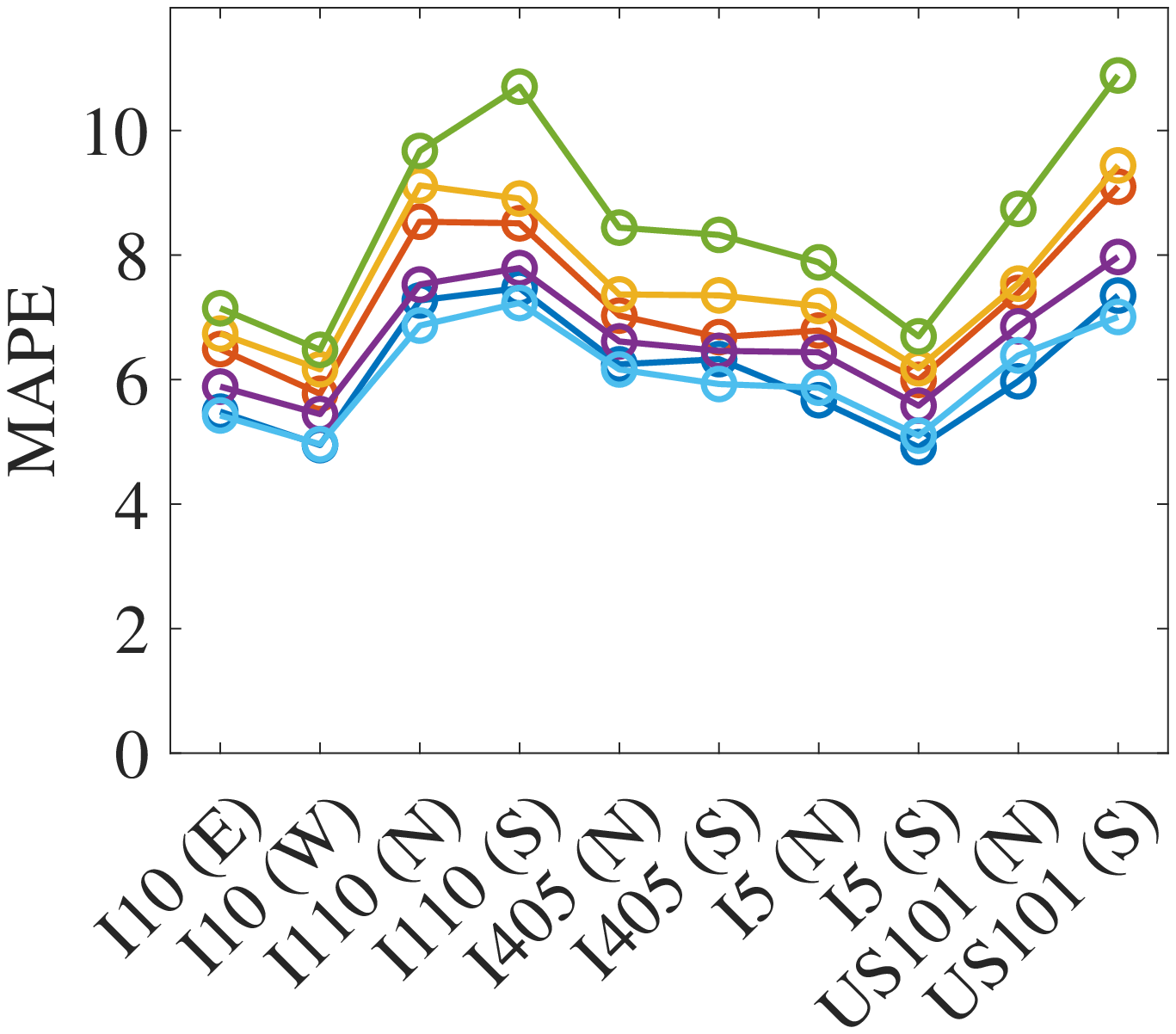}
    \caption{MAPE for prediction horizon of 15-min}\label{SubFig:MAPE_15}
  \end{subfigure}

  \begin{subfigure}[]{0.32\textwidth}
    \centering
    \includegraphics[width=\linewidth]{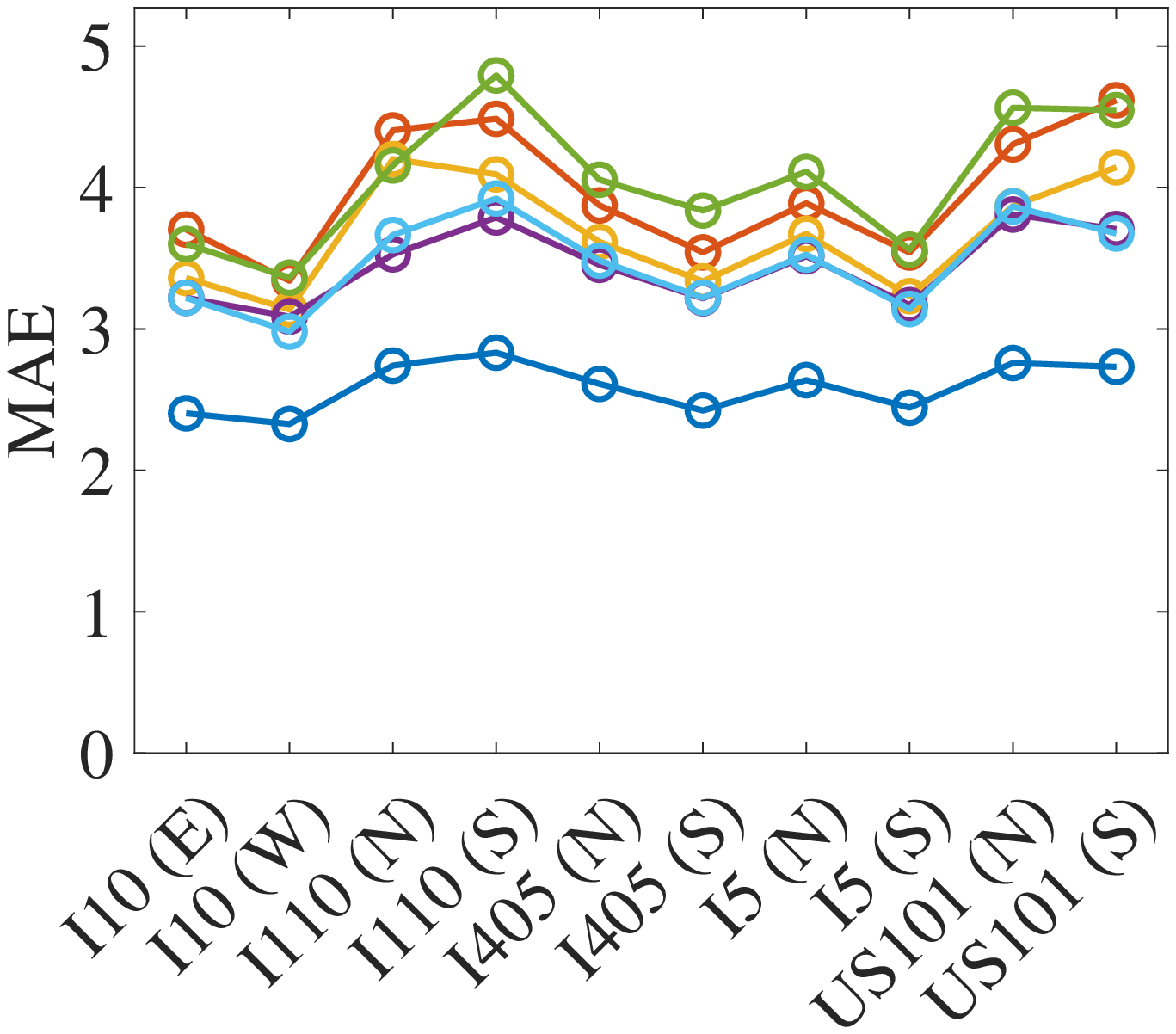}
    \caption{MAE for prediction horizon of 30-min}\label{SubFig:MAE_30}
  \end{subfigure}
  \hfill
  \begin{subfigure}[]{0.32\textwidth}
    \centering
    \includegraphics[width=\linewidth]{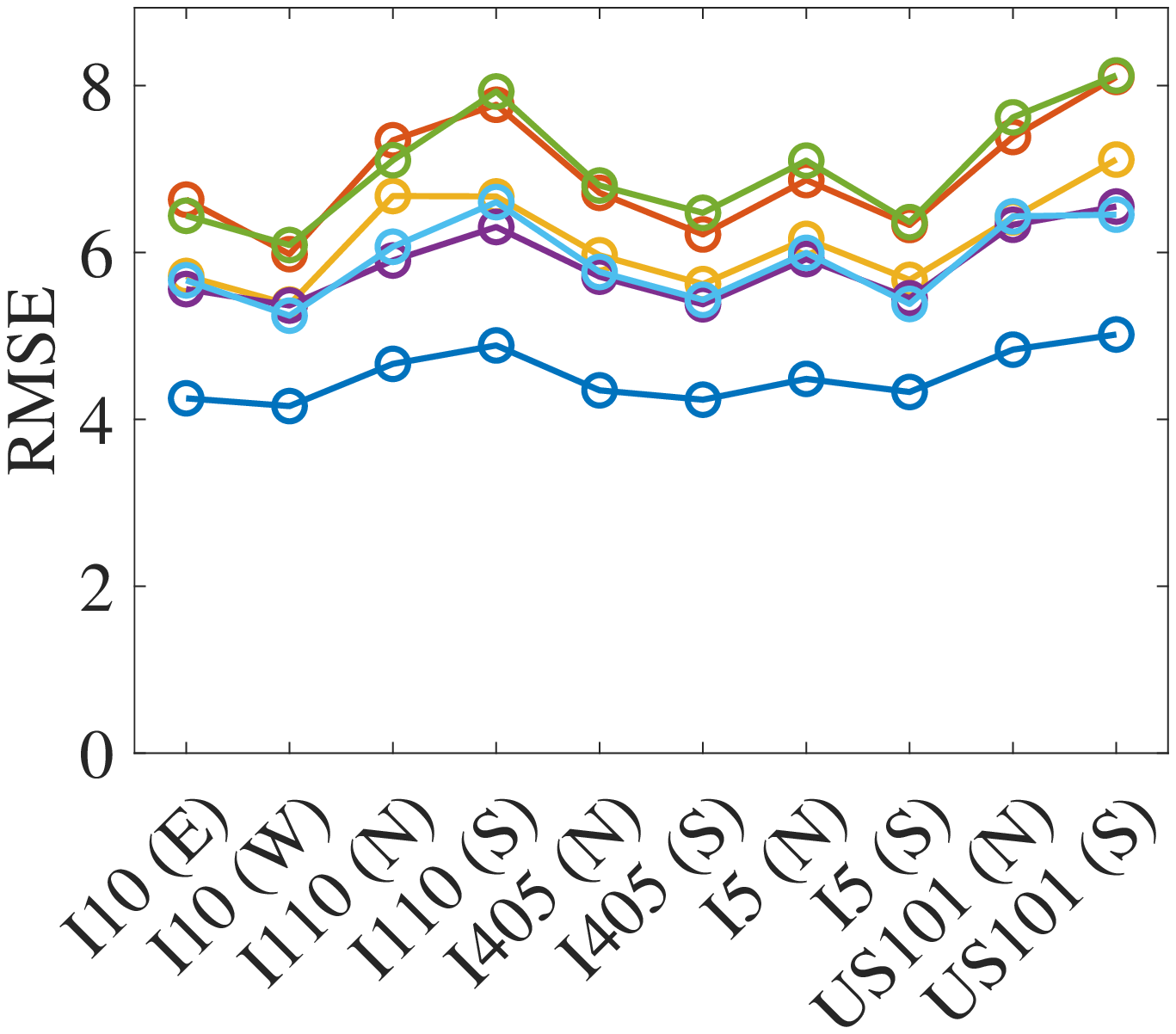}
    \caption{RMSE for prediction horizon of 30-min}\label{SubFig:RMSE_30}
  \end{subfigure}
  \hfill
  \begin{subfigure}[]{0.32\textwidth}
    \centering
    \includegraphics[width=\linewidth]{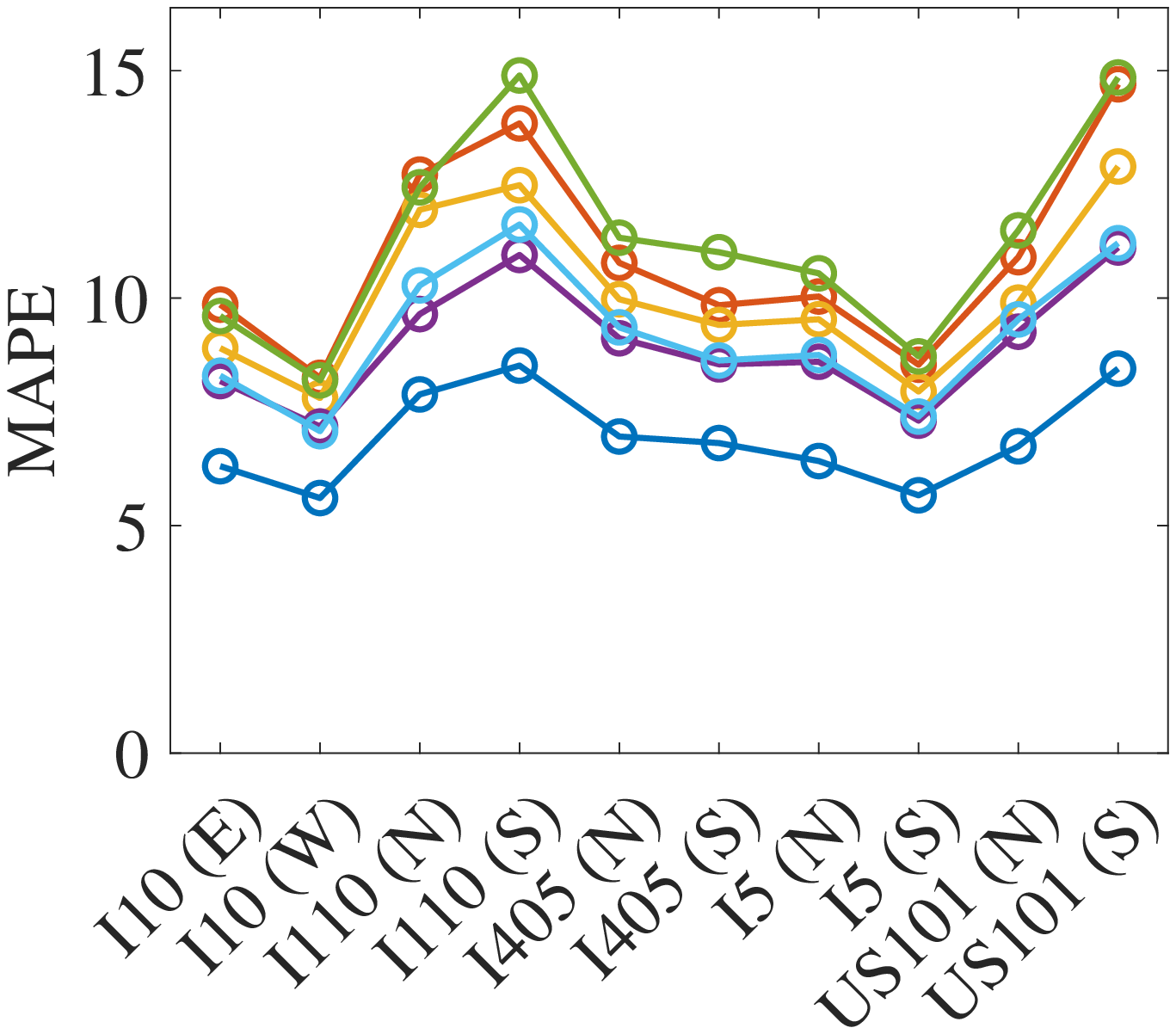}
    \caption{MAPE for prediction horizon of 30-min}\label{SubFig:MAPE_30}
  \end{subfigure}

  \begin{subfigure}[]{0.32\textwidth}
    \centering
    \includegraphics[width=\linewidth]{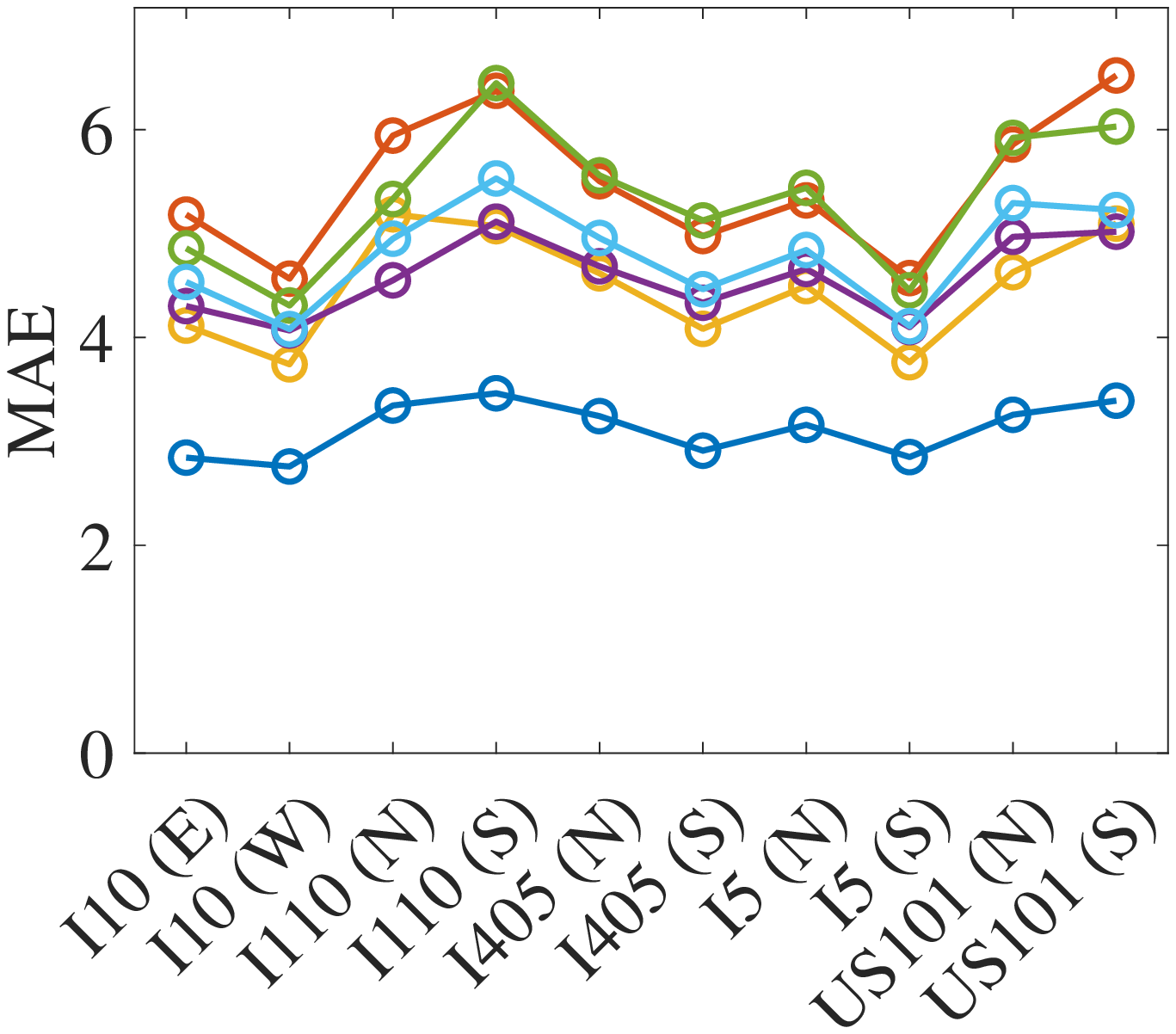}
    \caption{MAE for prediction horizon of 60-min}\label{SubFig:MAE_60}
  \end{subfigure}
  \hfill
  \begin{subfigure}[]{0.32\textwidth}
    \centering
    \includegraphics[width=\linewidth]{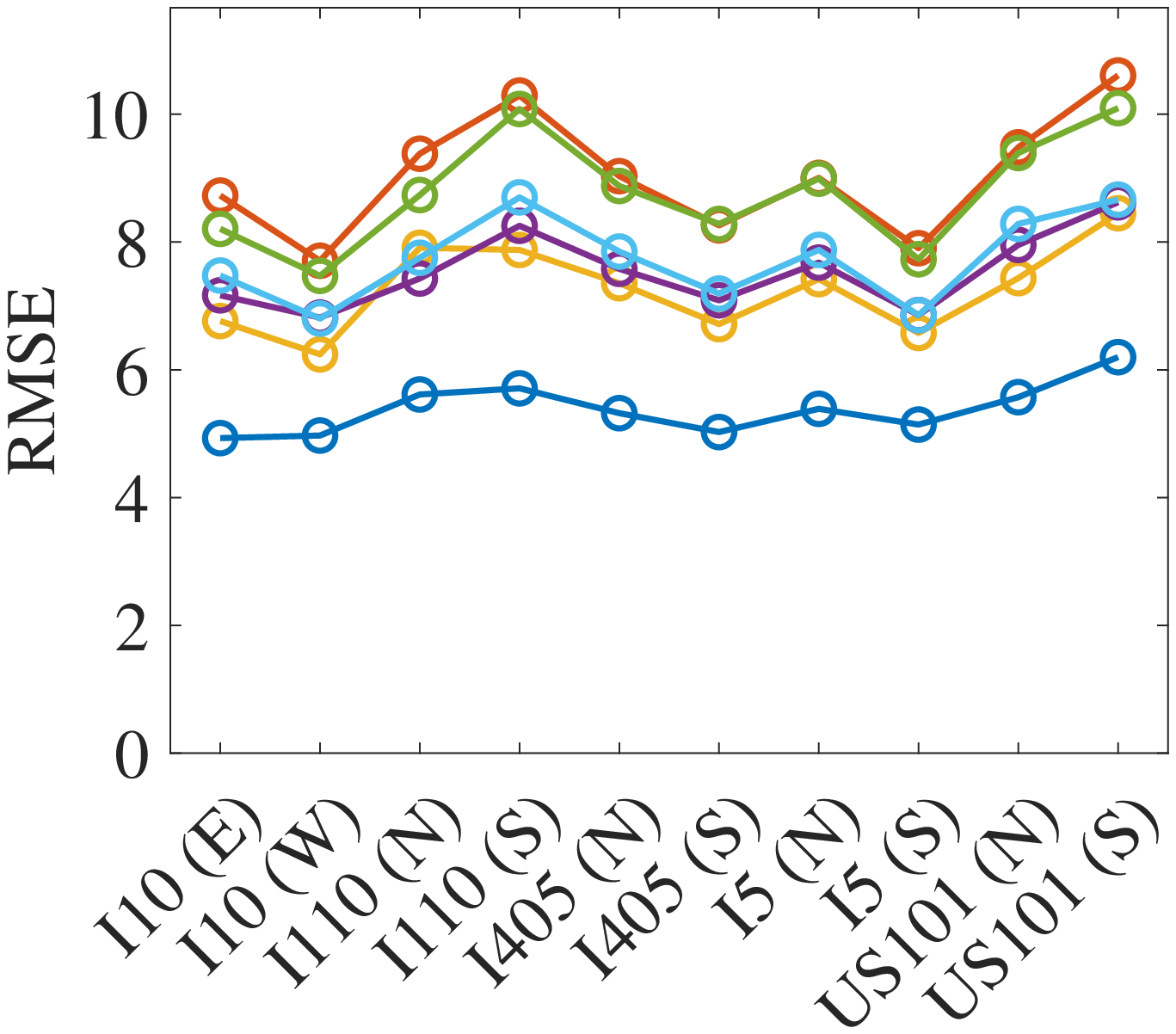}
    \caption{RMSE for prediction horizon of 60-min}\label{SubFig:RMSE_60}
  \end{subfigure}
  \hfill
  \begin{subfigure}[]{0.32\textwidth}
    \centering
    \includegraphics[width=\linewidth]{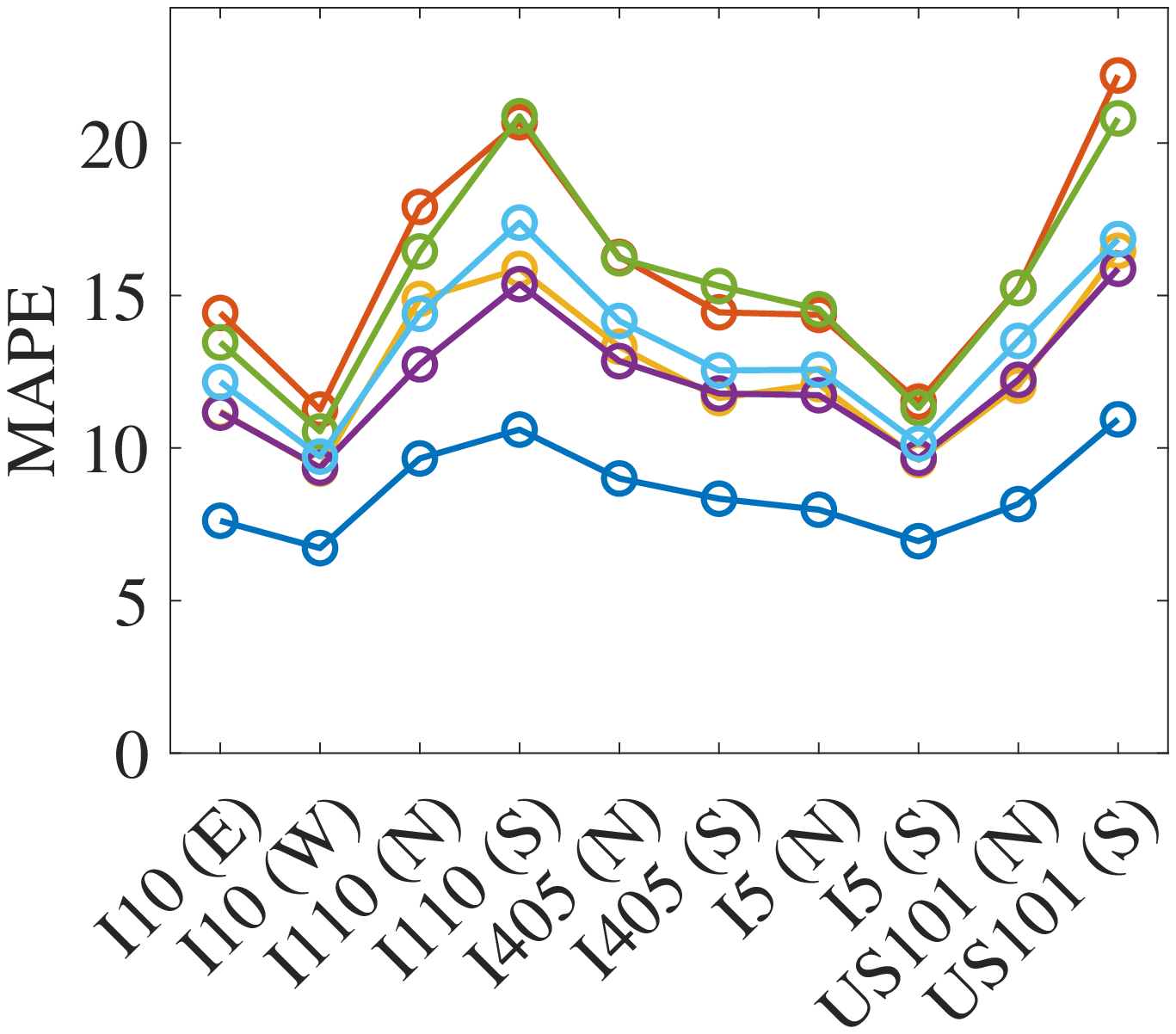}
    \caption{MAPE for prediction horizon of 60-min}\label{SubFig:MAPE_60}
  \end{subfigure}

  \caption{Performance comparison of the proposed approach with other techniques for each highway in the $PEMS-Los$ dataset. Legend: Proposed Approach (\protect\solidline[darkblue]), ANN (\protect\solidline[orange]), kNN (\protect\solidline[darkgreen]), XGBoost (\protect\solidline[skyblue]), CNN (\protect\solidline[lightorange]), LSTM (\protect\solidline[purple]).} \label{Fig:HighwayPerformanceComparison}
\end{figure}

The traffic speed prediction performance of the proposed approach at multiple time horizons for each highway of Los Angeles has also been compared with other techniques and the results are shown in Figure~\ref{Fig:HighwayPerformanceComparison}. There are a total of five highways in the $PEMS-Los$ dataset i.e. I5, I10, I110, I405, and US101. The prediction performance for both sides of traffic has been compared separately. In Figure~\ref{Fig:HighwayPerformanceComparison}, the letter in round brackets along with the highway name represents the traffic travel direction. A similar behavior, as depicted in Table \ref{Tab:Comparison}, can be observed in the figure. The proposed approach performed consistently well than other techniques on all highways of Los Angeles with the least error values. The consistency can also be observed by comparing errors with respect to time horizons. For instance, as shown in Figure~\ref{SubFig:MAPE_30} and \ref{SubFig:MAPE_60}, the MAPE of the proposed approach lies in the range of [5,10] approximately for both prediction time horizons of 30-min and 60-min whereas for the same time horizons the MAPE of other techniques has increased from [7,15] to [10,21] approximately.

\begin{figure}[!ht]
      \centering
      \includegraphics[scale=1]{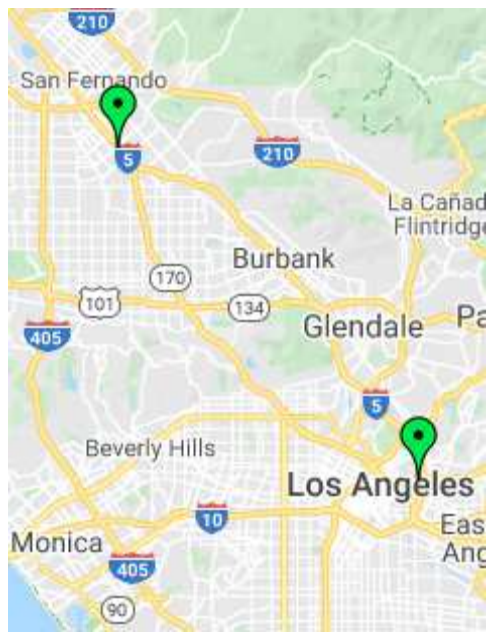}
      \caption{Location of sensors 716960 (top left) and 717040 (bottom right) on highway I5 and I10 in Los Angeles}\label{Fig:SensorLocation}
\end{figure}

\begin{figure}[H]
  \centering
  \begin{subfigure}[]{0.49\textwidth}
    \centering
    \includegraphics[width=\textwidth]{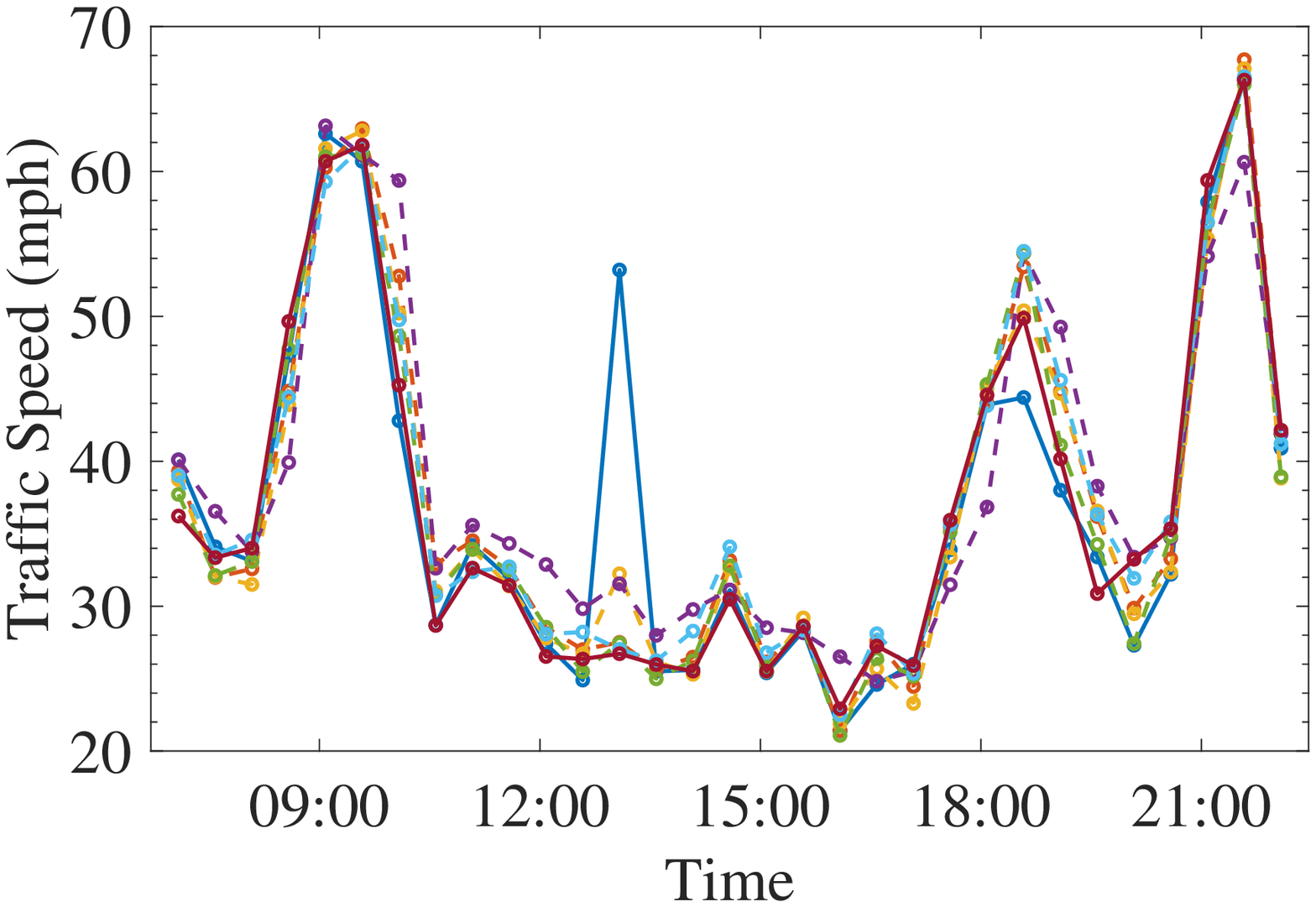}
    \caption{5-min traffic speed prediction for sensor 716960}\label{SubFig:S1_5}
  \end{subfigure}
  \hfill
  \begin{subfigure}[]{0.49\textwidth}
    \centering
    \includegraphics[width=\linewidth]{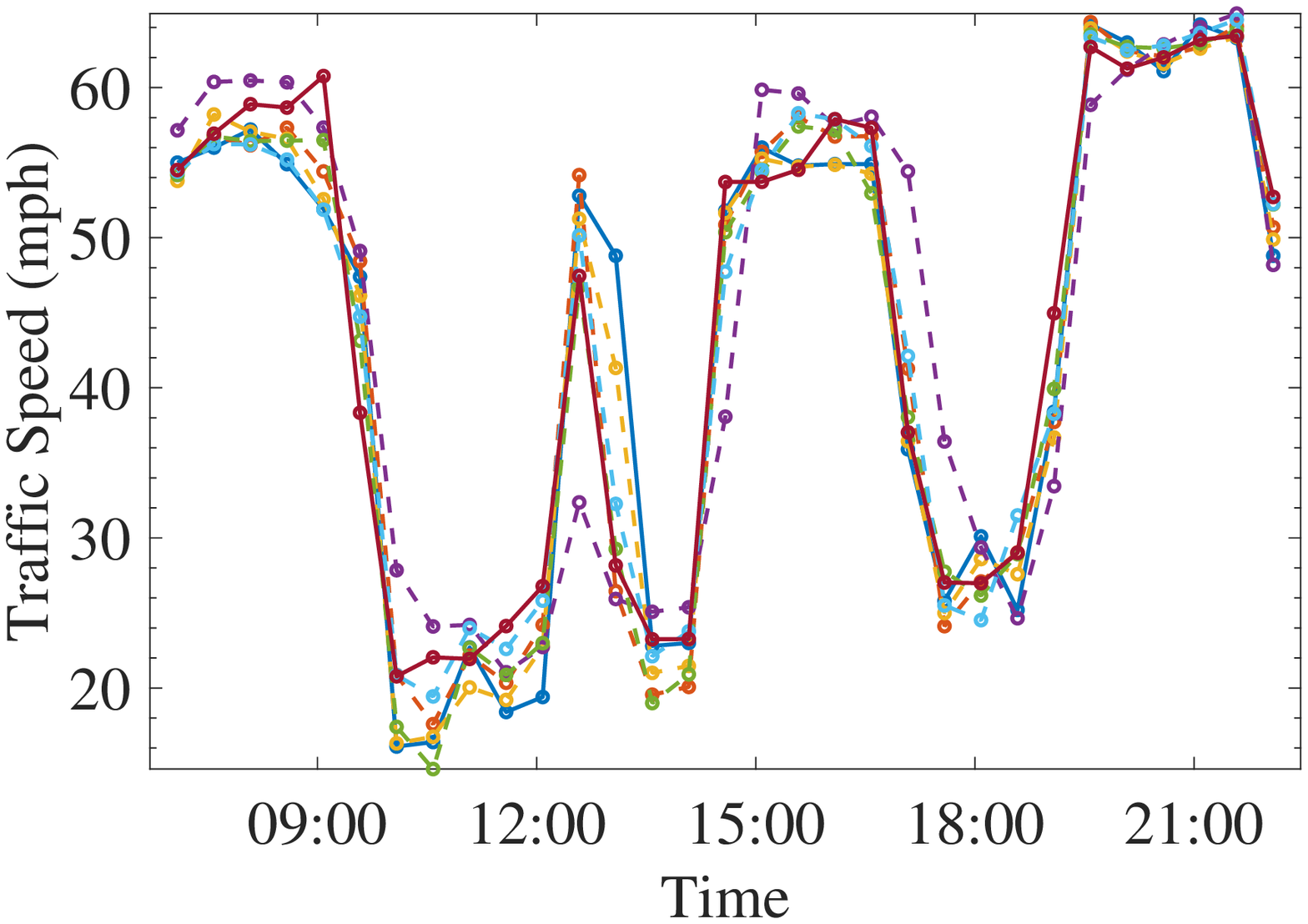}
    \caption{5-min traffic speed prediction for sensor 717040}\label{SubFig:S2_5}
  \end{subfigure}

  \begin{subfigure}[]{0.49\textwidth}
    \centering
    \includegraphics[width=\textwidth]{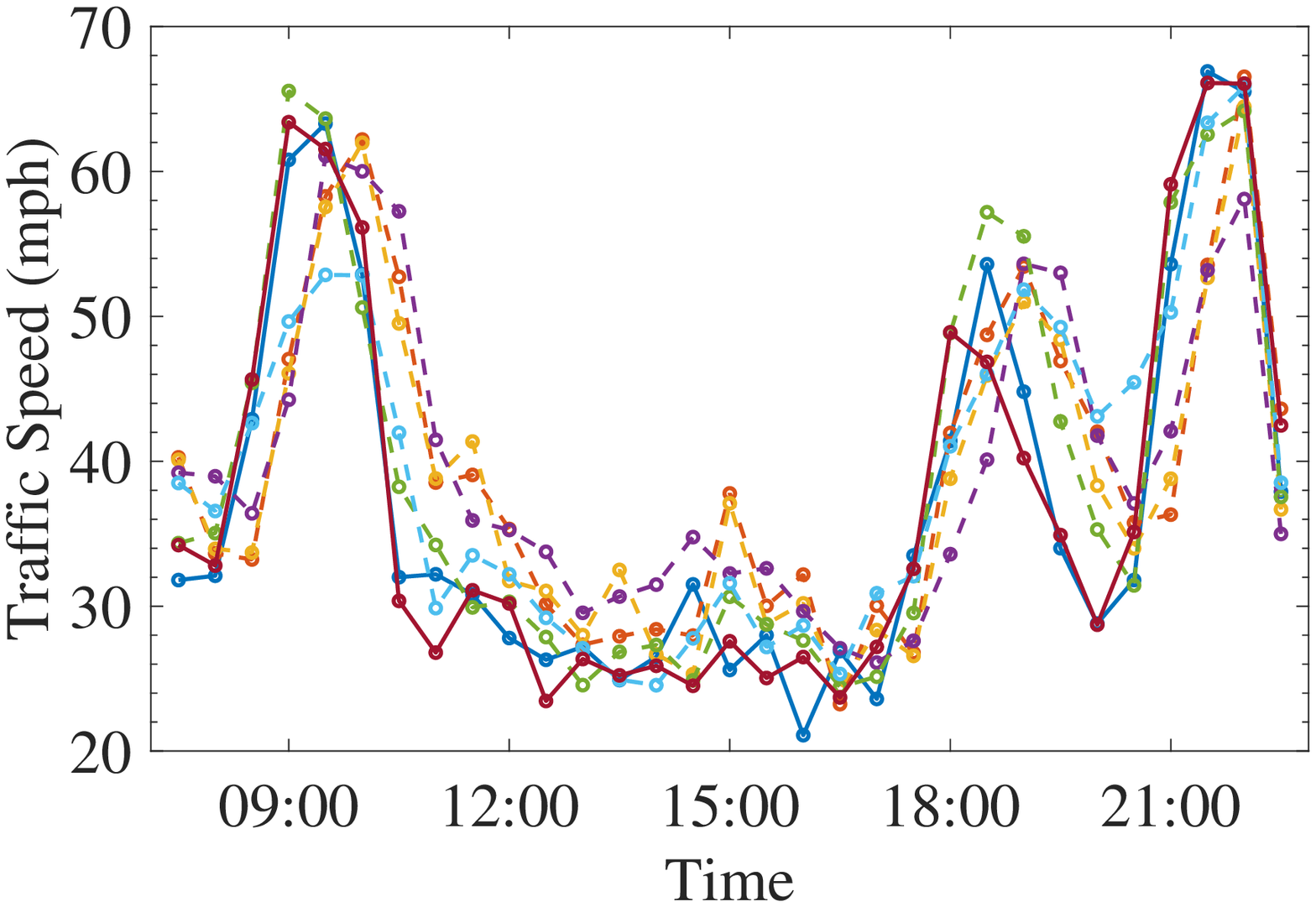}
    \caption{30-min traffic speed prediction for sensor 716960}\label{SubFig:S1_30}
  \end{subfigure}
  \hfill
  \begin{subfigure}[]{0.49\textwidth}
    \centering
    \includegraphics[width=\linewidth]{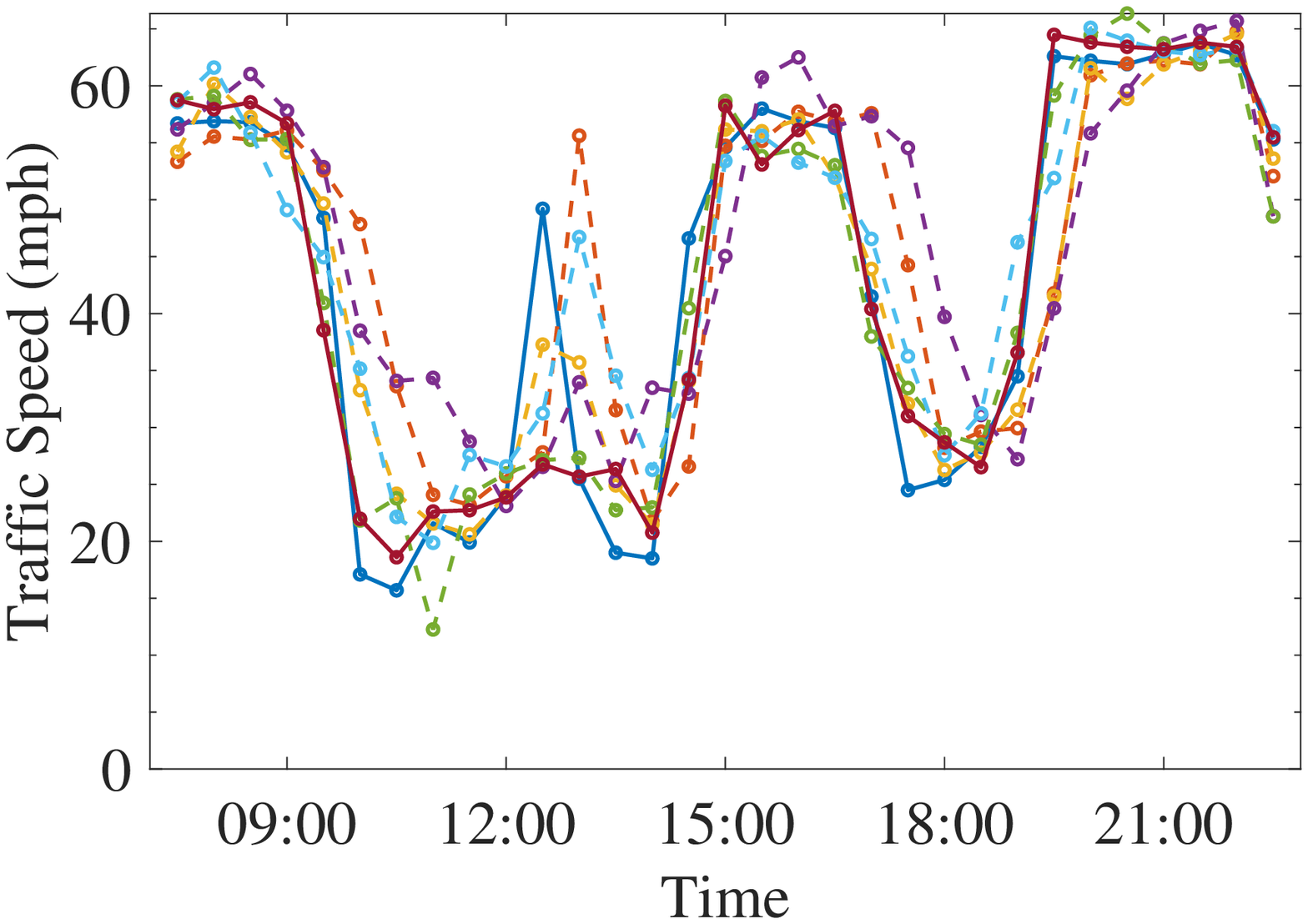}
    \caption{30-min traffic speed prediction for sensor 717040}\label{SubFig:S2_30}
  \end{subfigure}

  \begin{subfigure}[]{0.49\textwidth}
    \centering
    \includegraphics[width=\textwidth]{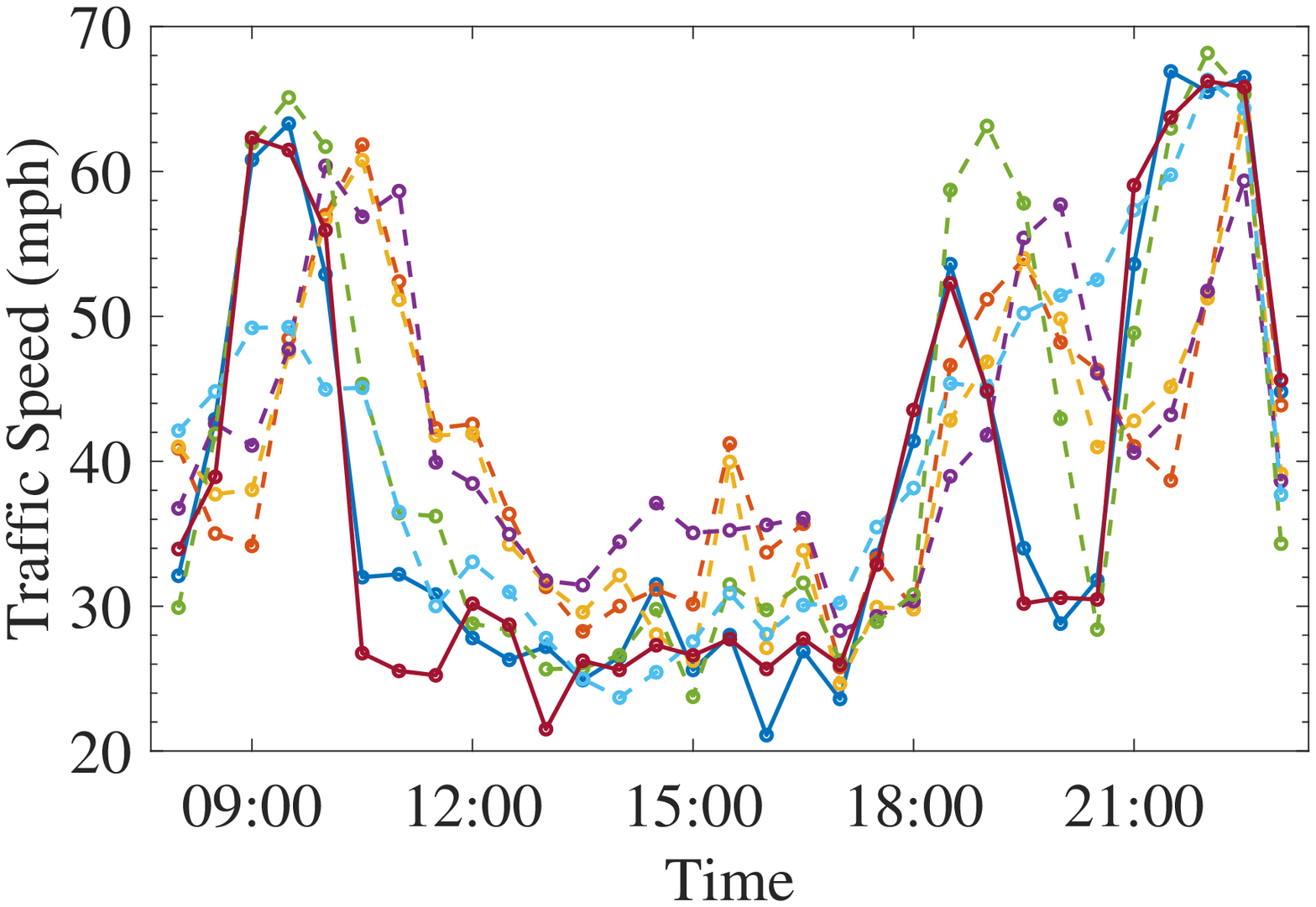}
    \caption{60-min traffic speed prediction for sensor 716960}\label{SubFig:S1_60}
  \end{subfigure}
  \hfill
  \begin{subfigure}[]{0.49\textwidth}
    \centering
    \includegraphics[width=\linewidth]{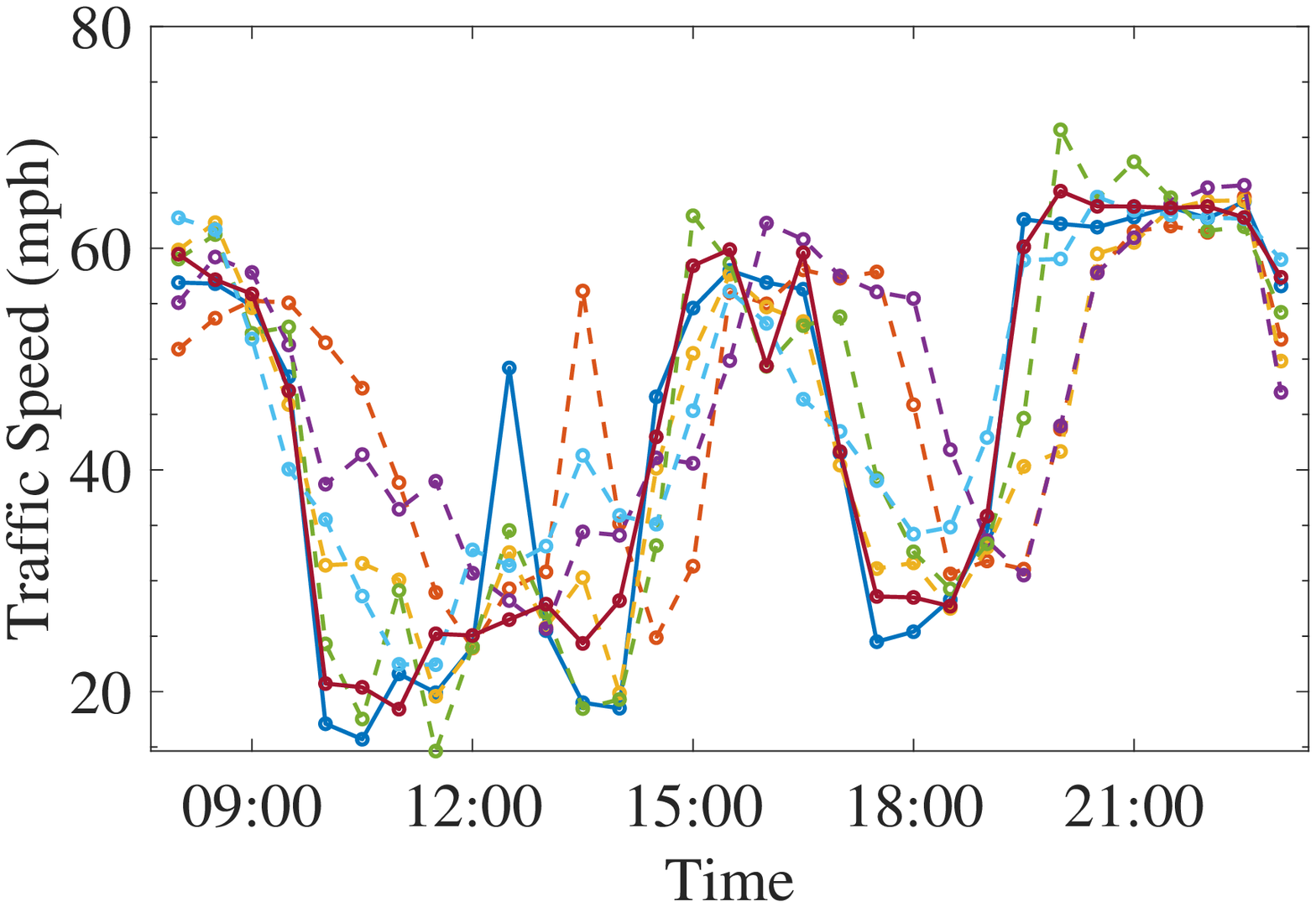}
    \caption{60-min traffic speed prediction for sensor 717040}\label{SubFig:S2_60}
  \end{subfigure}

  \caption{Prediction comparison on different time horizons of the proposed approach with other techniques for traffic data of 25th July, 2017 of two sensors 716960 and 717040 located in Los Angeles on highway I5 and I10, respectively. Legend: Actual (\protect\solidline[darkblue]), Proposed Approach (\protect\solidline[maroon]), ANN (\protect\dashedline[orange]), XGBoost (\protect\dashedline[lightorange]), kNN (\protect\dashedline[purple]), LSTM (\protect\dashedline[darkgreen]), CNN (\protect\dashedline[skyblue]).} \label{Fig:SensorWiseComparison}
\end{figure}

The difference in prediction performance of the proposed approach and other techniques is clearly visible in Figure~\ref{Fig:SensorWiseComparison}. The figure shows the comparison of one-day traffic prediction at three different time horizons (5-min, 30-min, and 60-min) by proposed and other benchmark techniques for two sensors 716960 and 717040 located on highways I5 and I10, respectively. The location of both the sensors on the map of Los Angeles has been shown in Figure~\ref{Fig:SensorLocation} for better understanding. In Figure~\ref{Fig:SensorWiseComparison}, it can be seen that the proposed technique consistently provides traffic speed prediction which is very near to the actual traffic speed. The difference is more clearly visible in Figures~\ref{SubFig:S1_60} and \ref{SubFig:S2_60}, where other techniques deviate from the actual traffic speed quite often while the proposed approach follows the pattern of actual traffic speed consistently. Also, it can be observed that the proposed technique is performing really well whenever there is a sudden transition in traffic speed, for instance, at around 9:00 and 18:00 hours (peak hours) there is a sudden change in traffic speed for both the sensors, as shown in Figures~\ref{SubFig:S1_30}, \ref{SubFig:S2_30}, \ref{SubFig:S1_60}, and \ref{SubFig:S2_60}, but the traffic prediction by proposed approach does not deviate much from the actual traffic speed as compared to other techniques.

From the above discussion, it can be concluded that the proposed technique provides considerably better traffic speed prediction results than the other state-of-the-art techniques. To further validate this conclusion two statistical tests namely, Kruskal Wallis H Test (KWT) and Multiple Comparison Test (MCT), have been performed. KWT is a rank-based non-parametric statistical test that internally uses one-way ANOVA. It tests the null hypothesis that the independent samples of two groups originated from the same distribution. On the other hand, the MCT determines which groups differ significantly by comparing several groups using one-way ANOVA. These tests were applied in sequential order. First, KWT was applied on MAE of prediction results obtained from proposed and other techniques and then, MCT was applied on the rank information provided by KWT.

The results obtained from KWT and MCT are shown in Figure~\ref{Fig:StatisticalResults}. In this figure, the left column shows the box and whisker plots for the results of KWT for traffic prediction on different time horizons. The boxes in each box plot have lines indicating $1^{st}$, $2^{nd}$, and $3^{rd}$ quartiles. The whiskers have vertical lines from the boxes to show the deviation of data. From the box and whisker plots, it can be concluded that the proposed approach has least MAE for all the time horizons except 5-min. This is in accordance with the results presented in Table~\ref{Tab:Comparison}. Also, from the size of the boxes, it can be inferred that the deviation in MAE is very less for the proposed approach as compared to other techniques in most cases. This shows the consistency and robustness of the proposed approach. The right column of Figure~\ref{Fig:StatisticalResults} shows the results obtained by performing MCT on different time horizons. In each plot of the right column, the proposed approach has been shown using a blue line. The techniques which have mean ranks significantly different from the proposed approach are shown with red lines. Also, techniques that do not differ significantly are shown using gray-colored lines. From these figures, it can be concluded that the proposed technique is significantly different from other techniques in most of the cases. There are overlaps of the proposed technique with ANN and LSTM in Figure~\ref{SubFig:MCT_5} and with XGBoost in Figure~\ref{SubFig:MCT_15}. It means that the proposed technique has comparable performance with these techniques on 5 and 15 min prediction horizon. However, for prediction the horizon of 15 min, the MAE of the proposed technique is less than XGBoost, so it can be concluded that the proposed technique provides better predictions. Also, for other prediction horizons, the proposed technique is better as there is no overlap and the proposed technique has least MAE and mean rank.

The statistical results of MCT have been given in Table~\ref{Tab:StatisticalTest}. In the table, the first column represents the prediction horizon, the second column provides the names of the techniques that are compared. The third, fourth, fifth, and sixth columns represent the lower bound, mean rank difference, upper bound, and p-value obtained by performing MCT for the pair of techniques, respectively. The lower and upper bound represent the minimum and maximum value of 95\% confidence interval, as the test was performed with significance level $\alpha=0.05$. From the table, it can be observed that the p-values for most of the pairs of techniques are less than the $\alpha$. This implies that for most of the pairs the techniques are significantly different and hence null hypothesis is rejected. This can also be observed from the lower and upper bound values. If the value of 0.0 does not lie within the confidence interval's limit then we can conclude that the techniques compared are significantly different at the significance level of 0.05. Also, as discussed above, the proposed approach is better due to least MAE for traffic prediction on all time horizons except on 5 min horizon. Even for prediction horizon of 5 min, the proposed technique provides comparable results.

\begin{figure}[H]
  \centering
  \begin{subfigure}[]{0.49\textwidth}
    \centering
    \includegraphics[width=\textwidth]{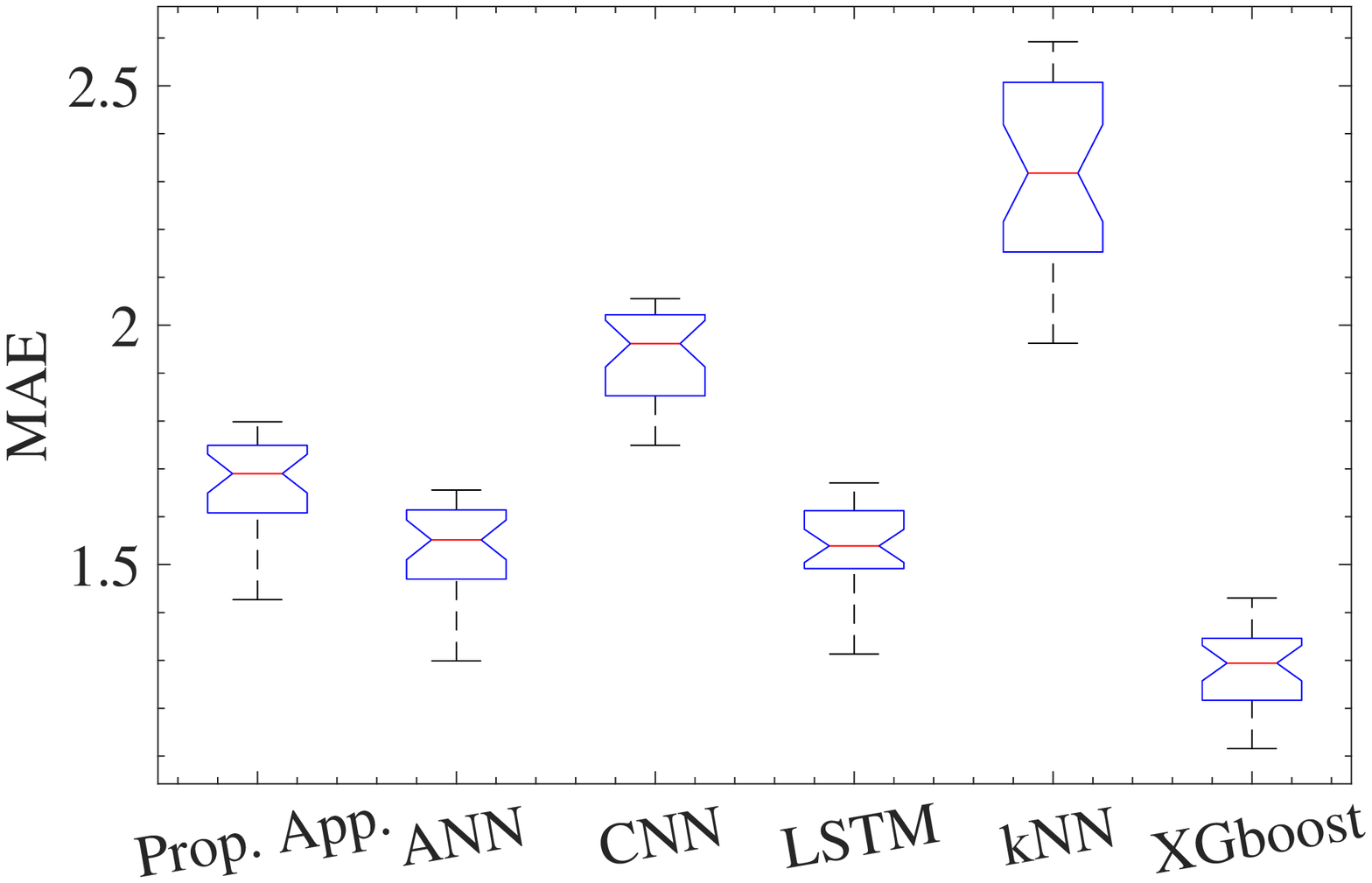}
    \caption{Kruskal Wallis Test on MAE for 5 min prediction}\label{SubFig:Kruskal_5}
  \end{subfigure}
  \hfill
  \begin{subfigure}[]{0.49\textwidth}
    \centering
    \includegraphics[width=\linewidth]{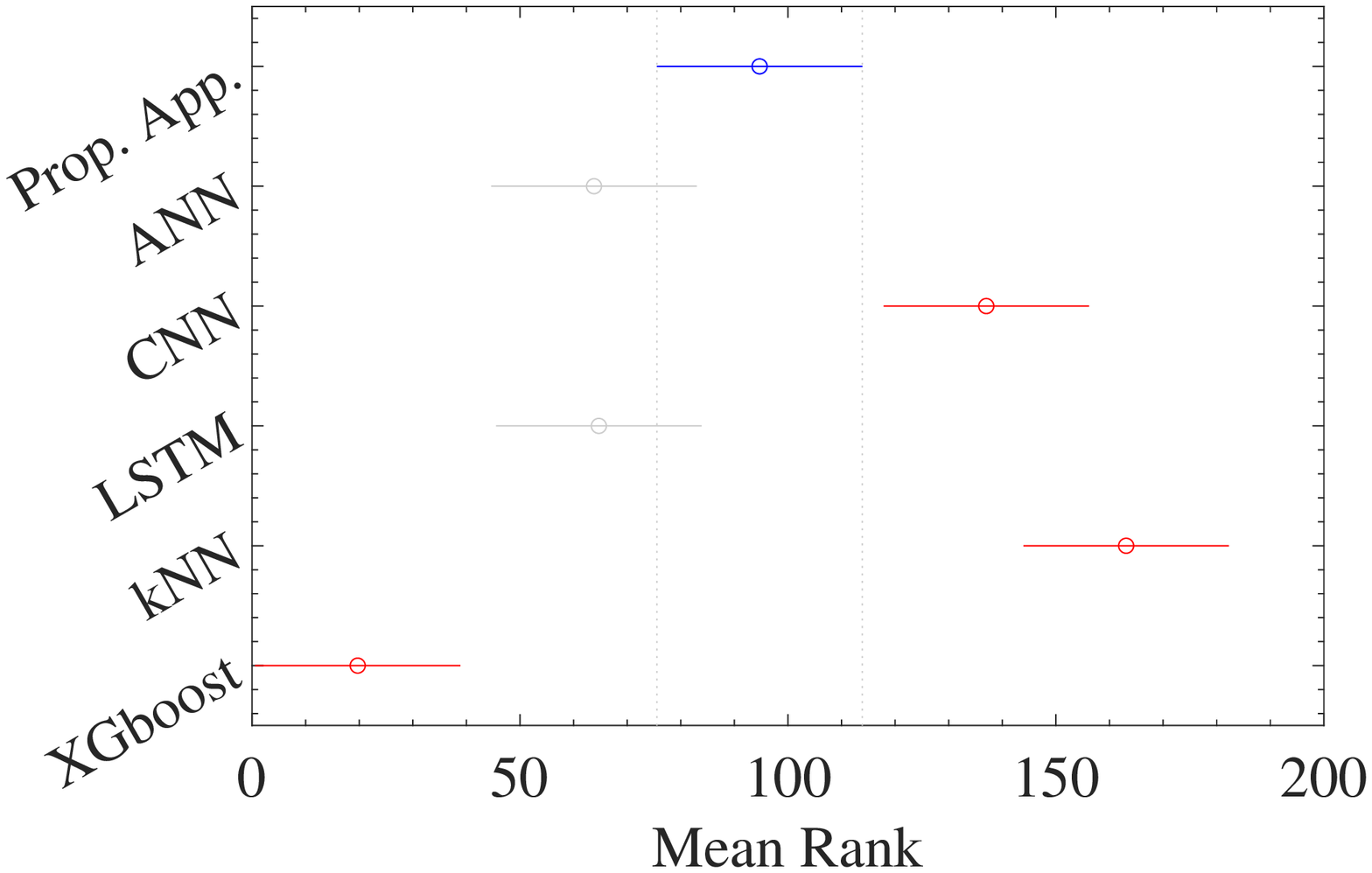}
    \caption{Multiple Comparison Test for 5 min prediction}\label{SubFig:MCT_5}
  \end{subfigure}

  \begin{subfigure}[]{0.49\textwidth}
    \centering
    \includegraphics[width=\textwidth]{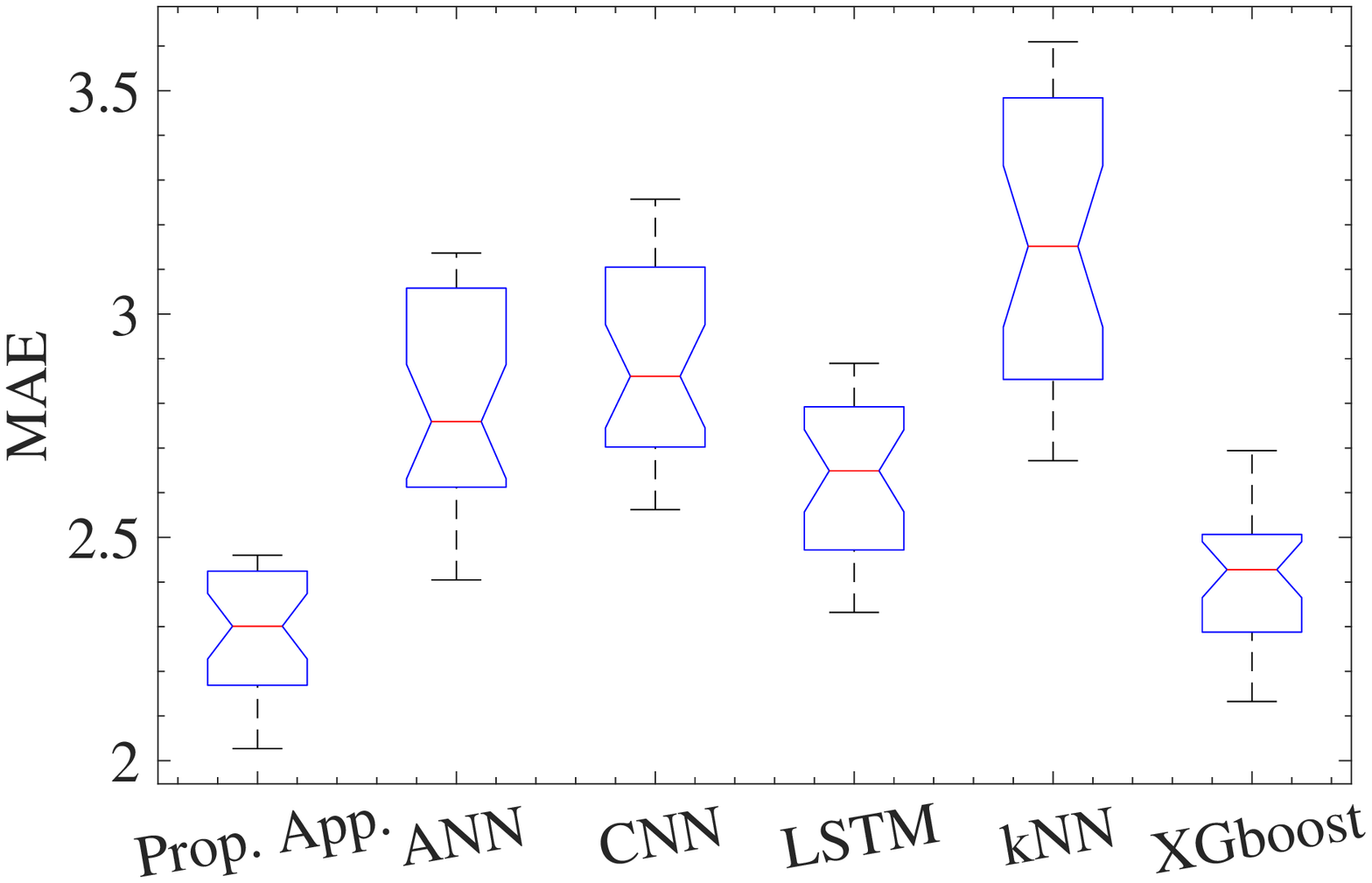}
    \caption{Kruskal Wallis Test on MAE for 15 min prediction}\label{SubFig:Kruskal_15}
  \end{subfigure}
  \hfill
  \begin{subfigure}[]{0.49\textwidth}
    \centering
    \includegraphics[width=\linewidth]{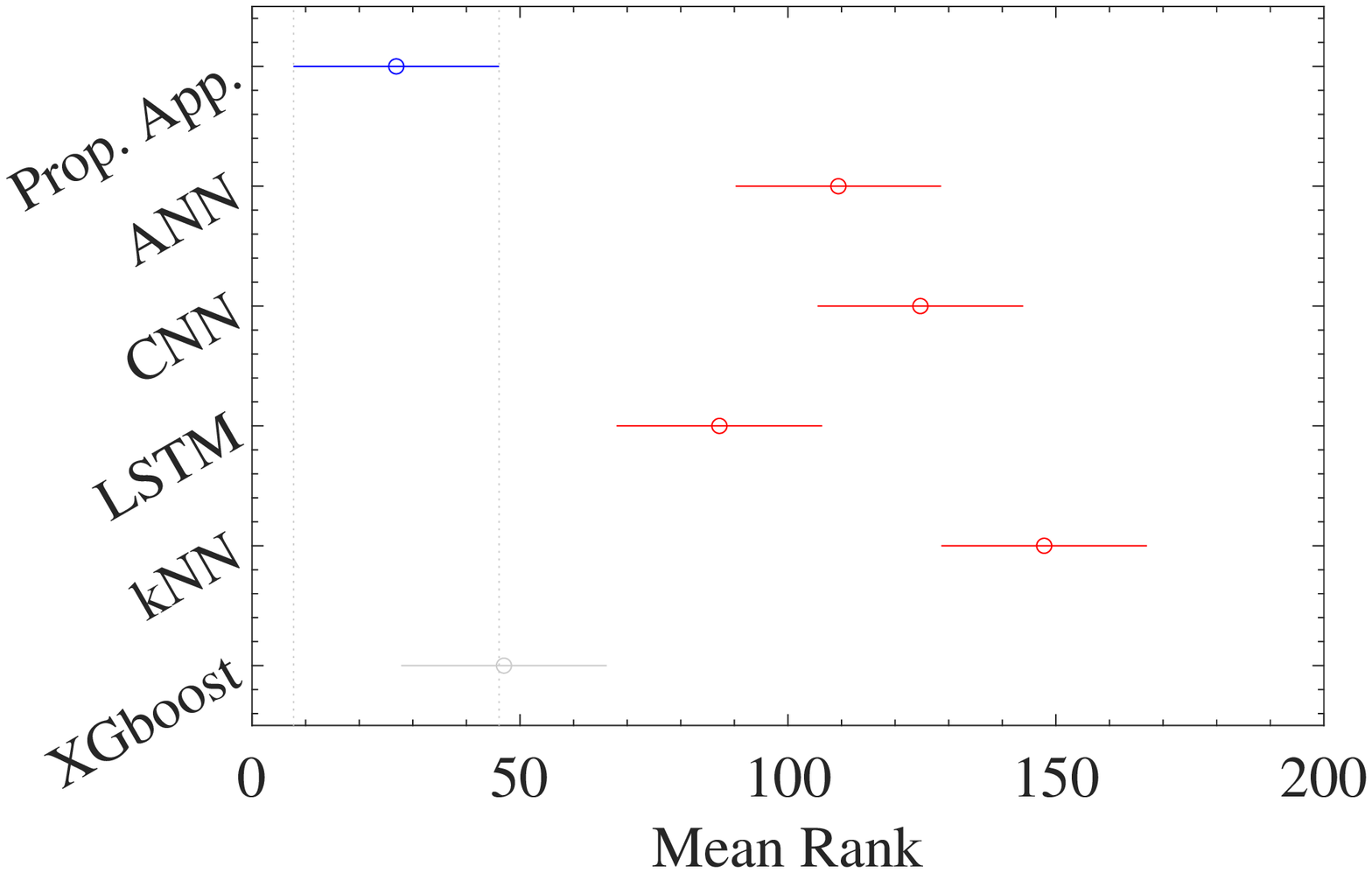}
    \caption{Multiple Comparison Test for 15 min prediction}\label{SubFig:MCT_15}
  \end{subfigure}

  \begin{subfigure}[]{0.49\textwidth}
    \centering
    \includegraphics[width=\textwidth]{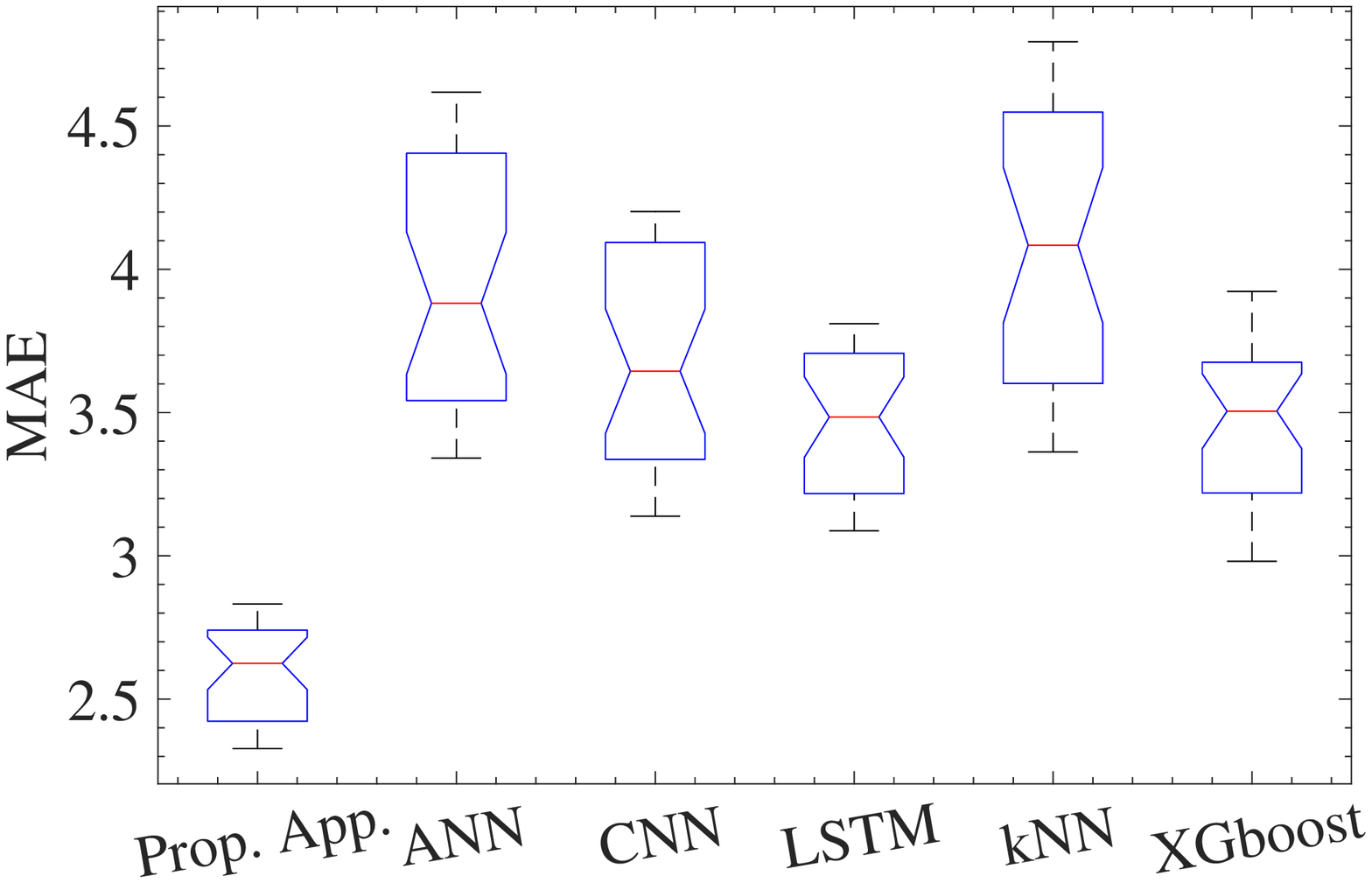}
    \caption{Kruskal Wallis Test on MAE for 30 min prediction}\label{SubFig:Kruskal_30}
  \end{subfigure}
  \hfill
  \begin{subfigure}[]{0.49\textwidth}
    \centering
    \includegraphics[width=\linewidth]{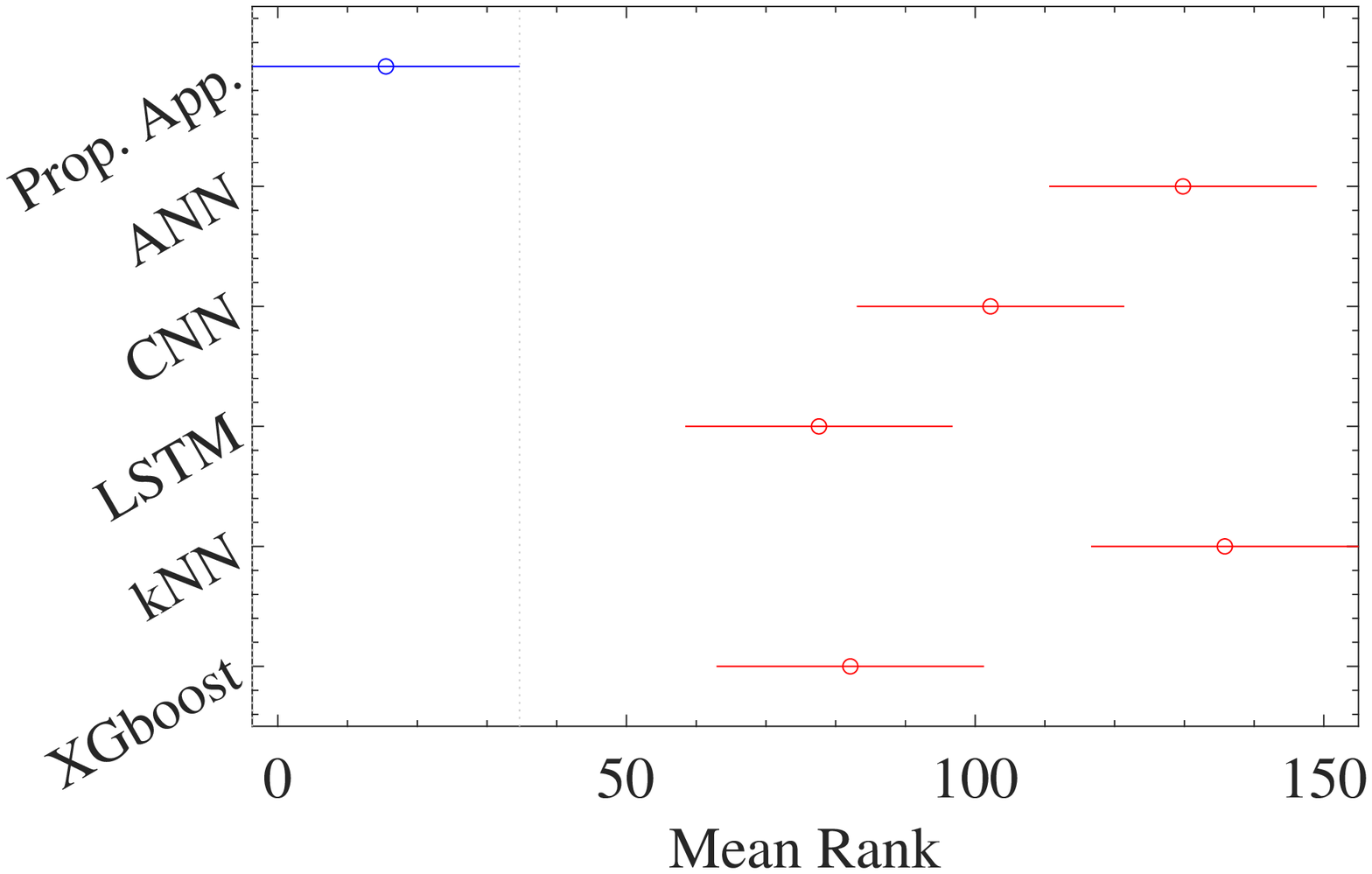}
    \caption{Multiple Comparison Test for 30 min prediction}\label{SubFig:MCT_30}
  \end{subfigure}

  \caption{Comparison results of predictions on different time horizons for two statistical tests namely, Kruskal Wallis and Multiple Comparison Test} 
\end{figure}

\begin{figure}[H]\ContinuedFloat
  \centering
  \begin{subfigure}[]{0.49\textwidth}
    \centering
    \includegraphics[width=\textwidth]{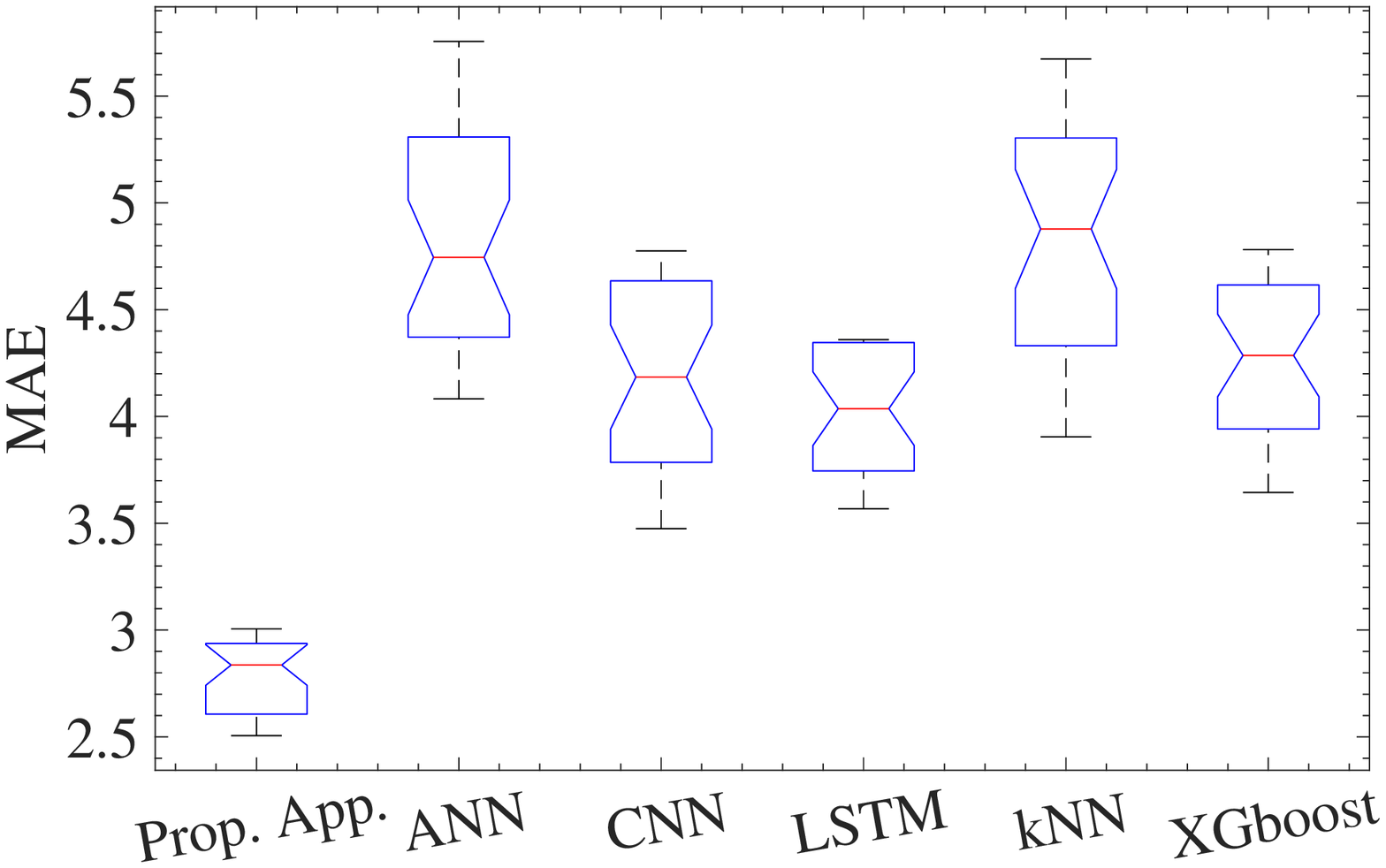}
    \caption{Kruskal Wallis Test on MAE for 45 min prediction}\label{SubFig:Kruskal_45}
  \end{subfigure}
  \hfill
  \begin{subfigure}[]{0.49\textwidth}
    \centering
    \includegraphics[width=\linewidth]{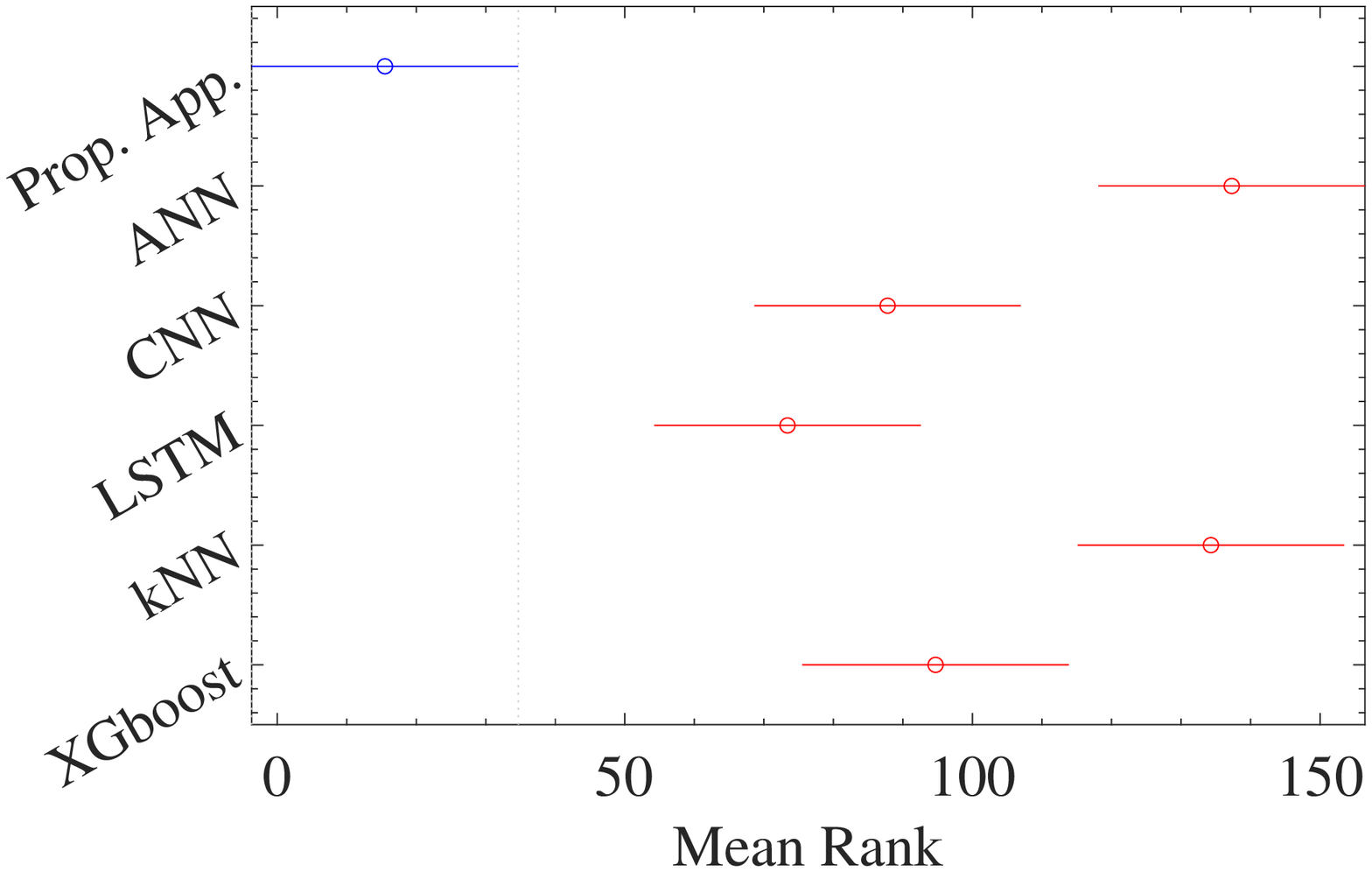}
    \caption{Multiple Comparison Test for 45 min prediction}\label{SubFig:MCT_45}
  \end{subfigure}

  \begin{subfigure}[]{0.49\textwidth}
    \centering
    \includegraphics[width=\textwidth]{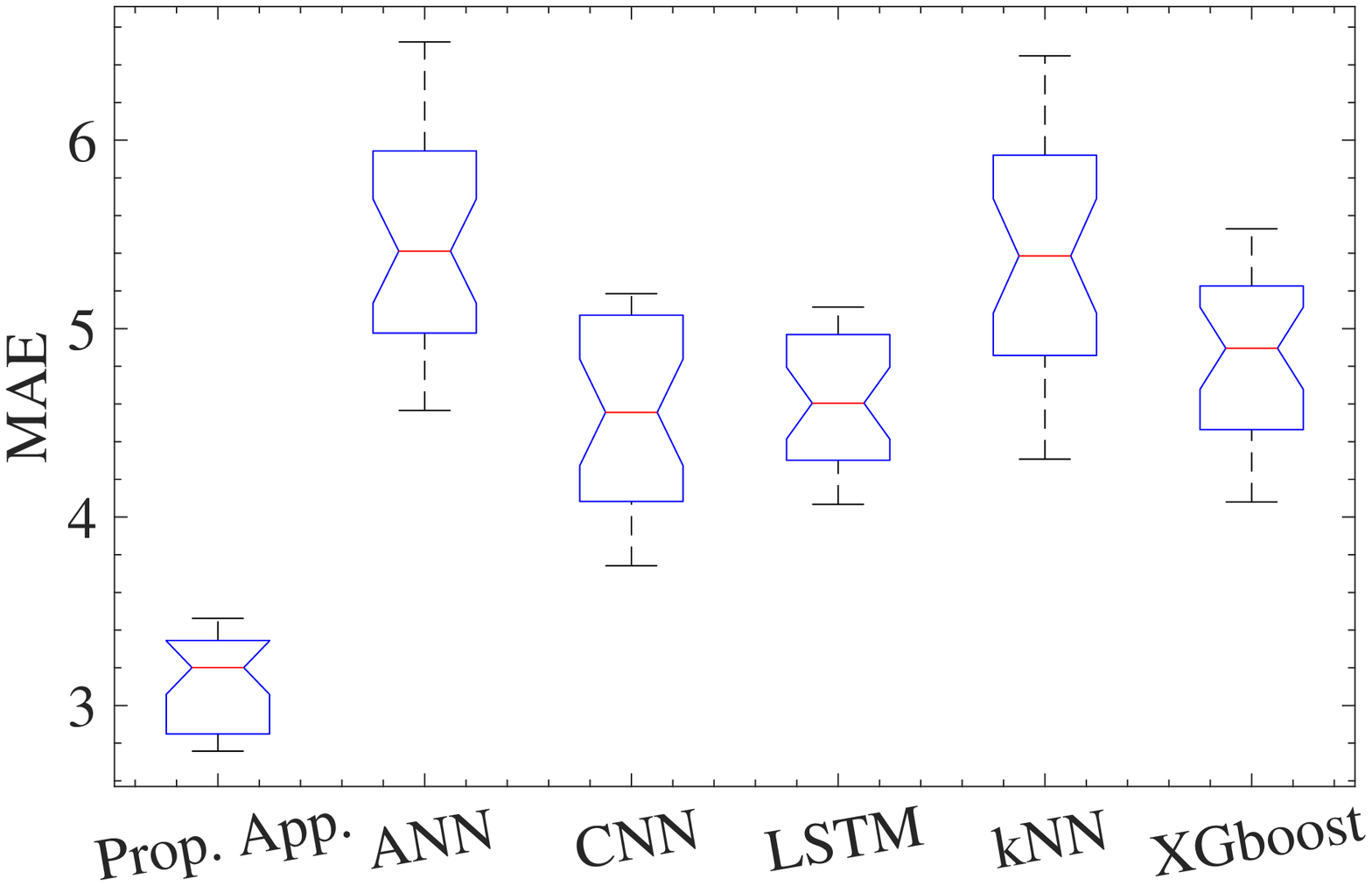}
    \caption{Kruskal Wallis Test on MAE for 60 min prediction}\label{SubFig:Kruskal_60}
  \end{subfigure}
  \hfill
  \begin{subfigure}[]{0.49\textwidth}
    \centering
    \includegraphics[width=\linewidth]{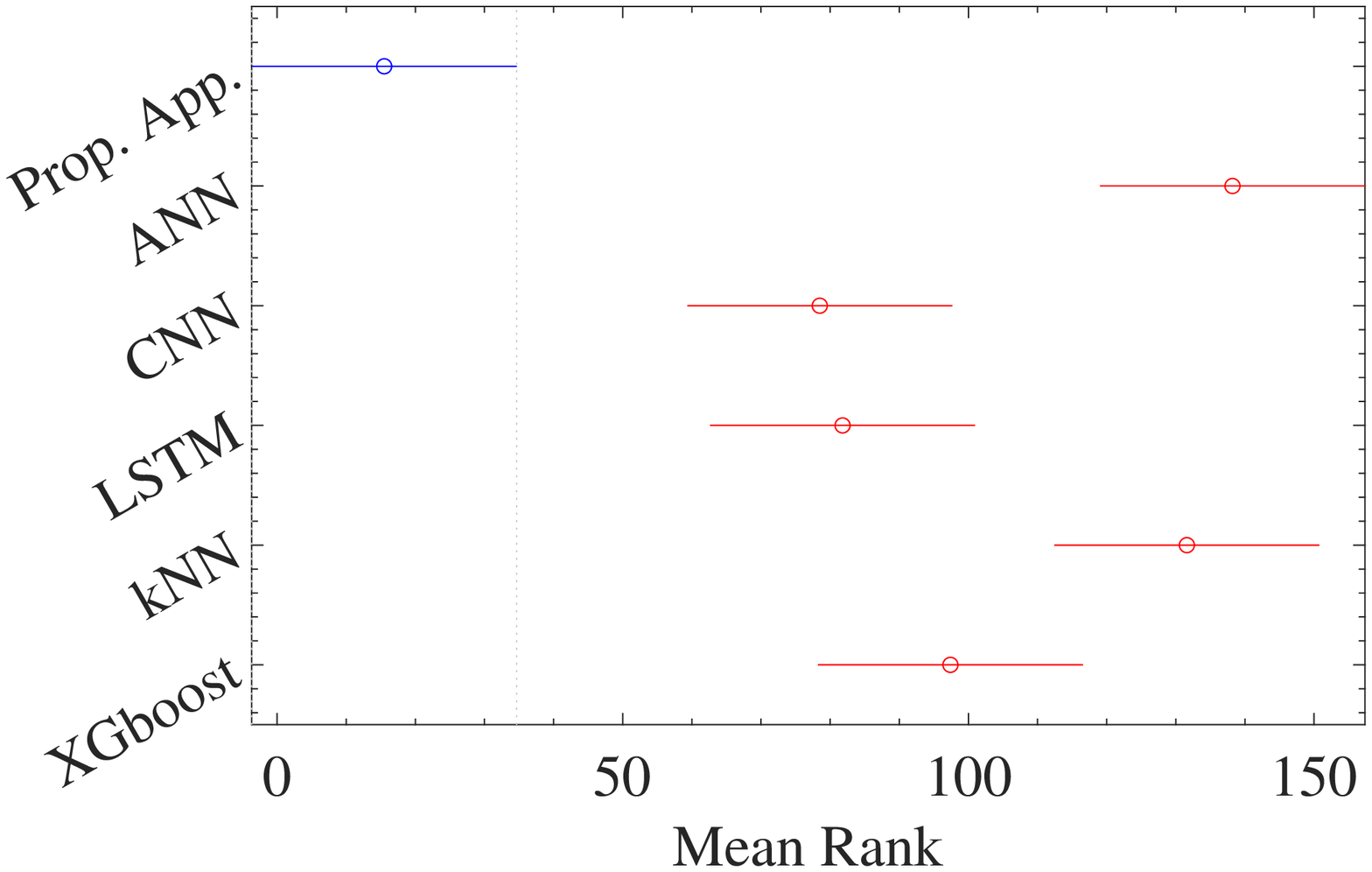}
    \caption{Multiple Comparison Test for 60 min prediction}\label{SubFig:MCT_60}
  \end{subfigure}

  \caption{Comparison results of predictions on different time horizons for two statistical tests namely, Kruskal Wallis and Multiple Comparison Test (Contd.)} \label{Fig:StatisticalResults}
\end{figure}

\begin{table}[h!]
    \centering
    \caption{Results of a statistical test using MCT}
    \label{Tab:StatisticalTest}
    \resizebox{\linewidth}{!}{

    \begin{tabular}{|c|c|C{3.7em}|C{5em}|C{3.7em}|c|}
    \hline
    Horizon                 & Comparison                   & Lower Bound   & Mean Rank Difference  & Upper Bound   & p-Value     \\ \hline

    \multirow{5}{*}{5 min}  & Proposed Approach vs ANN      & -7.434    & 30.9  & 69.234    & 1.95E-01  \\
                            & Proposed Approach vs CNN      & -80.634   & -42.3 & -3.966    & 2.06E-02  \\
                            & Proposed Approach vs LSTM     & -8.334    & 30.0  & 68.334    & 2.24E-01  \\
                            & Proposed Approach vs kNN      & -106.734  & -68.4 & -30.066   & 5.48E-06  \\
                            & Proposed Approach vs XGBoost  & 36.666    & 75.0  & 113.334   & 3.87E-07  \\ \hline
    \multirow{5}{*}{15 min} & Proposed Approach vs ANN      & -120.834  & -82.5 & -44.166   & 3.32E-08  \\
                            & Proposed Approach vs CNN      & -136.134  & -97.8 & -59.466   & 2.07E-08  \\
                            & Proposed Approach vs LSTM     & -98.634   & -60.3 & -21.966   & 1.08E-04  \\
                            & Proposed Approach vs kNN      & -159.234  & -120.9& -82.566   & 2.07E-08  \\
                            & Proposed Approach vs XGBoost  & -58.434   & -20.1 & 18.234    & 6.68E-01  \\ \hline
    \multirow{5}{*}{30 min} & Proposed Approach vs ANN      & -152.634  & -114.3& -75.966   & 2.07E-08  \\
                            & Proposed Approach vs CNN      & -125.034  & -86.7 & -48.366   & 2.23E-08  \\
                            & Proposed Approach vs LSTM     & -100.434  & -62.1 & -23.766   & 5.73E-05  \\
                            & Proposed Approach vs kNN      & -158.634  & -120.3& -81.966   & 2.07E-08  \\
                            & Proposed Approach vs XGBoost  & -104.934  & -66.6 & -28.266   & 1.10E-05  \\ \hline
    \multirow{5}{*}{45 min} & Proposed Approach vs ANN      & -160.134  & -121.8& -83.466   & 2.07E-08  \\
                            & Proposed Approach vs CNN      & -110.634  & -72.3 & -33.966   & 1.16E-06  \\
                            & Proposed Approach vs LSTM     & -96.234   & -57.9 & -19.566   & 2.42E-04  \\
                            & Proposed Approach vs kNN      & -157.134  & -118.8& -80.466   & 2.07E-08  \\
                            & Proposed Approach vs XGBoost  & -117.534  & -79.2 & -40.866   & 7.83E-08  \\ \hline
    \multirow{5}{*}{60 min} & Proposed Approach vs ANN      & -161.034  & -122.7& -84.366   & 2.07E-08  \\
                            & Proposed Approach vs CNN      & -101.334  & -63.0 & -24.666   & 4.15E-05  \\
                            & Proposed Approach vs LSTM     & -104.634  & -66.3 & -27.966   & 1.23E-05  \\
                            & Proposed Approach vs kNN      & -154.434  & -116.1& -77.766   & 2.07E-08  \\
                            & Proposed Approach vs XGBoost  & -120.234  & -81.9 & -43.566   & 3.73E-08  \\ \hline
\end{tabular}}

\end{table}

All of these techniques were tested on a system with Intel Xeon Processor and 12 GB RAM. In terms of inference time, ANN is the fastest by giving prediction results in just 8.87E-4 sec. XGBoost comes next with the inference time of 7.98E-3 sec. After XGBoost, three techniques namely, CNN, LSTM, and the proposed method have best inference time of 5.47E-2, 9.26E-2 and 9.69E-2 sec, respectively. kNN is the slowest with the prediction time of 1.03E-1 sec. So, it can be observed that the proposed method has comparable inference time with CNN and LSTM but is slower than ANN and XGBoost. This is justifiable as there are a large number of computation layers in the proposed model's architecture and hence more computation is required. The computation time can be further reduced by using GPUs or converting the proposed model to TensorFlow Lite format. Although the computation time is higher for proposed technique, the prediction accuracy is also higher than other techniques as discussed previously.

To generalize the effectiveness of the proposed approach on traffic data from different locations, the proposed technique was trained and tested using traffic data from $PEMS-Bay$. Also, comparison results with other state-of-the-art techniques that are already implemented on this dataset have been presented in Table~\ref{Tab:ComparisonPEMS-Bay}. The table presents a comparison of the proposed technique with four existing techniques namely, Diffusion Convolutional Recurrent Neural Network (DCRNN) (\cite{Li2017}), Spatio-Temporal Graph Convolutional Network (STGCN) (\cite{Learning2018}), Graph Wavenet (\cite{ijcai2019-264}) and Graph Multi-Attention Network (GMAN) (\cite{Zheng2020}). All of these techniques are graph-based techniques that use a time sequence of node-edge graphs (representing the road network traffic) as input to predict future traffic.

\begin{table}[h]
    \centering
    \caption{Comparison of the proposed approach with other popular approaches on PEMS-Bay dataset}
    \label{Tab:ComparisonPEMS-Bay}

    \begin{tabular}{|c|c|c|c|c|}
        \hline
        Technique               & Metric         & 15-min    & 30-min       & 60-min            \\ \hline

        \multirow{3}{*}{GMAN}   & MAE            & 1.34      & 1.62         & 1.86          \\
                                & RMSE           & 2.82      & 3.72         & 4.32          \\
                                & MAPE           & 2.81      & 3.63         & 4.31          \\ \hline

        \multirow{3}{*}{Graph Wavenet}
                                & MAE            & \textbf{1.30}      & 1.63         & 1.95          \\
                                & RMSE           & 2.74      & 3.70         & 4.52          \\
                                & MAPE           & \textbf{2.73}      & 3.67         & 4.63          \\ \hline

        \multirow{3}{*}{DCRNN}  & MAE            & 1.38      & 1.74         & 2.07          \\
                                & RMSE           & 2.95      & 3.97         & 4.74          \\
                                & MAPE           & 2.90      & 3.90         & 4.90          \\ \hline

        \multirow{3}{*}{STGCN}  & MAE            & 1.36      & 1.81         & 2.49          \\
                                & RMSE           & 2.96      & 4.27         & 5.69          \\
                                & MAPE           & 2.90      & 4.17         & 5.79          \\ \hline

        \multirow{3}{*}{Proposed Approach}  & MAE   & 1.38 & \textbf{1.49} & \textbf{1.75}     \\
                                            & RMSE  & \textbf{2.30} & \textbf{2.68} & \textbf{3.12}     \\
                                            & MAPE  & 3.02 & \textbf{3.22} & \textbf{3.97}     \\ \hline
        \multicolumn{5}{|l|}{\footnotesize{Values in bold represent the best ones}}                             \\ \hline
    \end{tabular}

\end{table}

From the results presented in Table~\ref{Tab:ComparisonPEMS-Bay}, it can be observed that the technique proposed in this work performs consistently better than the above discussed graph-based techniques, especially for longer prediction horizons like 30-min and 60-min. The main reason for this improved performance is the capability of the proposed technique to better understand the long-term traffic patterns. All of these graph-based techniques considered only short-term historical traffic data (mostly past 1 hour) as input whereas the proposed technique takes historical traffic data even from previous and previous to previous week. This helps the proposed technique to understand the weekly traffic pattern in a more effective manner. Also, the proposed neighbor sensor selection algorithm helps in selecting only those neighbor sensors which are spatially relevant. This helps the model to focus only on the most relevant information. So, from the above discussion, it can be concluded that the proposed technique provides better traffic speed prediction by considering both spatial and temporal traffic dependencies.

\section{Conclusion and Future Work}\label{Sec:Conclusion}
A deep learning based approach has been developed which can provide accurate multistep traffic speed prediction by considering the spatio-temporal traffic dependency. The proposed approach was trained and tested with the real-world traffic speed data collected from a number of loop detector sensors located on different highways of Los Angeles and California's Bay Area. The spatio-temporal dependencies were taken into account by selecting the nearby sensors based on traffic similarity and distance. The selection changes for each time instance. The proposed approach contains two deep auto-encoders that were cross-connected using the concept of latent space mapping. The cross-connected auto-encoders were then trained using the historical traffic data from the selected nearby sensors. The effectiveness of the proposed technique is validated by comparing it with multiple state-of-the-art techniques. The comparison results confirm that the proposed technique provides better prediction accuracy with the least errors at different prediction time horizons. For longer prediction horizon like 30 to 60 min, where other techniques provide traffic prediction with more error, the proposed technique still provide accurate predictions. This proves the consistency and reliability of the proposed technique. The proposed approach was trained and tested on traffic data from two different locations and the results of traffic prediction show the better performance of proposed approach. This shows the generalizability of the proposed approach.

In the future, the proposed technique can be further extended to predict long-term traffic speed like next-day traffic speed prediction. Although the weekends and weekdays both are considered for traffic prediction, it will be interesting to consider the traffic pattern for holidays other than weekends. Also, even though the proposed approach considers the effect of environmental conditions or social events implicitly by neighboring sensor selection algorithm, the prediction accuracy can be improved by considering these factors explicitly.




\end{document}